\documentclass[aapm,mph,preprint,graphicx]{revtex4-1}

\usepackage{graphicx}
\usepackage{multirow}
\usepackage{amsfonts,amsmath, amsbsy, amssymb}
\usepackage{epsfig}
\usepackage{hyperref}
\usepackage[displaymath,mathlines]{lineno}
\usepackage{subfigure}
\usepackage{multirow}
\usepackage{amsbsy}
\usepackage{amsfonts}
\usepackage{amsmath}
\usepackage{amssymb}

\newcommand{\ab}{{\mathbf a}}

\newcommand{\eb}{{\mathbf e}}
\newcommand{\fb}{{\mathbf f}}
\newcommand{\gb}{{\mathbf g}}

\newcommand{\pb}{{\mathbf p}}

\newcommand{\xb}{{\mathbf x}}
\newcommand{\yb}{{\mathbf y}}

\newcommand{\Sd}{{\mathbb S}}

\newcommand{\Hc}{{\mathcal H}}
\newcommand{\Ic}{{\mathcal I}}

\newcommand{\Mc}{{\mathcal M}}
\newcommand{\Nc}{{\mathcal N}}

\newcommand{\Qc}{{\mathcal Q}}
\newcommand{\Rc}{{\mathcal R}}
\newcommand{\Sc}{{\mathcal S}}
\newcommand{\Tc}{{\mathcal T}}

\newcommand{\thetab}{{\boldsymbol{\theta}}}

\newcommand{\nub}{{\boldsymbol{\nu}}}

\newcommand{\Rd}{{\mathbb R}}

\DeclareMathOperator{\sgn}{sgn}

\modulolinenumbers[5]

\begin{document}

\title{\Large One Network to Solve All ROIs:  Deep Learning  CT  for Any ROI using Differentiated Backprojection}

\author{Yoseob Han, and Jong Chul Ye$^*$}

\affiliation{BISPL - Bio Imaging, Signal Processing, and Learning lab. \\ Dept. of Bio and Brain Engineering, KAIST \\Daejeon, Republic of Korea\\
			 $^{*}$Corresponding author: jong.ye@kaist.ac.kr}

\begin{abstract}

\textbf{Purpose:}
Computed tomography for the reconstruction of region of interest (ROI) has advantages in reducing the X-ray dose and the use of a small detector.  However, standard analytic reconstruction methods such as filtered back projection (FBP) suffer from severe cupping artifacts, and existing model-based iterative reconstruction methods  require extensive computations. Recently, we proposed a  deep neural network to learn  the cupping artifacts, but the network was not  generalized well for different ROIs due to the singularities in the corrupted images. Therefore, there is an increasing demand for a neural network that works well for any ROI sizes.\par

\noindent\textbf{Method:}
Two types of neural networks are designed. The first type  learns  ROI size-specific cupping artifacts from FBP images, whereas the second type network is  for the  inversion of the truncated Hilbert transform from the truncated differentiated backprojection (DBP) data. Their generalizabilities for different ROI sizes, pixel sizes, detector pitch and starting angles for short-scan  are then investigated.

\noindent\textbf{Results:}
Experimental results show that the new type of neural networks significantly outperform existing iterative methods for all ROI sizes despite significantly lower runtime complexity. In addition, performance improvement is consistent across different acquisition scenarios.

\noindent\textbf{Conclusions:}
Since the proposed method consistently surpasses existing methods, it can be used as a general CT reconstruction engine for many practical applications without compromising possible detector truncation.

\end{abstract}

\maketitle


\section{Introduction}
X-ray computed tomography (CT) is one of the most powerful clinical imaging tools, delivering high-quality images in a fast and cost effective manner.
However,  X-ray radiation from CT increases the potential risk of cancers to patients,
so many studies has been conducted to reduce the X-ray dose.
In particular, low-dose X-ray CT technology has been extensively developed by reducing the number of photons, projection views, or ROI sizes.
Among these, the interior tomography aims to obtain an ROI image by irradiating only within the ROI.
Interior tomography is useful when the ROI within a patient's body is small (such as heart).
In some applications, interior tomography  has additional benefits 
thanks to the cost saving from the use of a small-sized detector.
However, 
the use of an analytic CT reconstruction algorithm generally produces images with severe cupping artifacts due to the transverse directional
projection truncation.

Sinogram extrapolation is a simple but inaccurate approximation method to reduce the artifacts \cite{hsieh2004algorithm}.
Recently, Courdurier et al \cite{courdurier2008solving} found that if an intensity of any ROI subarea is known {\em a priori}, the ROI area can be uniquely reconstructed from the cropped sinogram.
Assuming some prior knowledges of the functional space for images,
Katsevich et al \cite{katsevich2012stability} demonstrated the mathematical uniqueness of the interior tomography and provided an estimate of stability.
In Jin et al \cite{jin2012interior},
a continuous domain singular value decomposition (SVD) was performed on a truncated Hilbert transform operator to represent an ROI images with a span of the eigen-functions of the truncated Hilbert transform operator described in Katsevich et al \cite{katsevich2012finite}.  Then, they compensated for the null space using a general prior information.
Yu et al \cite{yu2009compressed} showed that an ROI images can be uniquely reconstructed  using the total variation (TV) penalty,  if an images is composed of piecewise smooth areas.
In a series of papers \cite{ward2015interior,lee2015interior}, our group has also shown that it is possible to reconstruct each chord line through the ROI using a generalized L-spline penalty \cite{ward2015interior};
and we have further confirmed that  the high frequency signal can be  analytically recovered  thanks to the Bedrosian identity,  whereas the computationally intensive iterative routine only need to be performed to recover the low frequency signal after downsampling \cite{lee2015interior}.
While this approach significantly reduces the computational complexity of the interior reconstruction,
the computational complexity of this technique, as well as most existing iterative reconstruction algorithms, still prohibits their routine clinical use.

In recent years, deep learning algorithms 
have achieved significant success in various
applications \cite{krizhevsky2012imagenet,ronneberger2015u,kang2017deep, chen2017low,han2016deep,jin2017deep}.
In particular,  various deep learning architecture 
have been successfully used for low-dose CT \cite{kang2017deep, kang2018deep,chen2017low}, sparse view CT \cite{han2016deep,jin2017deep,han2018framing}, etc.
These deep learning applications surpassed the previous iterative methods in image quality and reconstruction time.
Moreover, in recent theoretical works \cite{ye2018deep,ye2019cnn},
the authors showed that
deep learning is closely related to  a novel signal representation using non-local basis convolved with data-driven local basis,  
which make it useful for inverse problems.

Inspired by these findings, here we propose  deep learning frameworks for interior tomography problem.
One  important contribution of this paper is the observation that there are two ways of addressing the interior tomography, which can be directly
translated into two  distinct neural network architectures. 
More specifically, it is well-known that the technical difficulties of interior tomography arises from  the null space in the truncated Hilbert transform\cite{katsevich2012finite}.
One way to address this difficulty is a post-processing approach to remove the null space image from the analytic reconstruction.
In fact, our preliminary work  \cite{han2018roi} is the realization
of such idea in neural network domain, which was trained to  learn the cupping artifacts corresponding to the null space images.
On the other hand, a direct inversion can be done from the truncated DBP data using an inversion formula for finite Hilbert transform \cite{king2009hilbert}.
While this approach has been investigated by several pioneering works for interior tomography problems \cite{defrise2006truncated},
the main limitation of these approaches is that  
the selection of the optimal parameter to compensate for the null space   is difficult.
Therefore, another novel contribution is the second type of  neural network
that is designed to  invert the finite Hilbert transform from the truncated DBP data 
by learning the null space parameters and convolutional kernel for Hilbert transform from the training data.
%

Although the two neural network approaches appear similar except their inputs, there are fundamental differences in their generalization capability.
The first type network  learns the null space images from the artifacts corrupted input images. 
Although the approach \cite{han2018roi} provides near-perfect reconstruction with about $7\sim 10$ dB improvement in PSNR over existing methods \cite{yu2009compressed, lee2015interior},
the null space component of the analytic reconstruction contains the singularity at the ROI boundary with strong intensity saturation, so the trained
network for particular ROI size does not generalize well under different conditions. 
On the other hand, the input image for the second type network is the truncated DBP images, which correspond to the full DBP images on an ROI mask. Therefore, there are no singularities in the DBP images, which makes the network  generalize better  for different ROI sizes.
Numerical results showed that  while 
the second type network outperforms the existing interior tomography techniques for all ROIs in terms of image quality and reconstruction time, the first type network degrades  if the test phase imaging condition differs from the training data.
Moreover,  we demonstrate that the second type neural network provides consistent performance improvement for different acquisition parameters, e.g. different pixel resolution, detector pitch, and starting angle for the short-scan,  which are different from those at the training phase.
All these results confirmed the generalizability  of the new network architecture. 

This paper is structured as follows. In Section~\ref{sec:theory}, 
the basic theory of differentiated backprojection (DBP) and Hilbert transform are reviewed, the interior tomography problem is formally defined,
 from which
two types of neural network architectures are derived.
Then, Section~\ref{sec:method} describes the methods to implement and validate the proposed method,
which is followed by experimental results in Section~\ref{sec:result}.
Conclusions are provided in  Section~\ref{sec:conclusion}.

\section{Theory}
\label{sec:theory}


\subsection{Differentiated Backprojection and Hilbert Transform}

Let $\thetab$ denote a vector on the unit sphere $\Sd\in \Rd^3$.  The
set of vectors that are orthogonal to $\thetab$ is described as
$$\thetab^\perp=\{\yb \in \Rd^3~:~\yb^T\thetab = 0 \},$$
where $^T$ is the transpose.
The Radon transform of an image  $f(\xb)$, $\xb \in  \Rd^3$ is represented as
\begin{eqnarray}
\Rc f(\thetab,s):= \int_{\thetab^\perp} f(s\thetab+\yb) d\yb
\end{eqnarray}
where $s\in \Rd$ and $\thetab \in \Sd$. 
The X-ray transform $D_f$, which maps a function on $\Rd^3$ into the set of its line integrals, is defined as
\begin{equation}\label{eq:D_f}
D_f(\ab,\thetab)=\int_0^\infty dt~f(\ab+t\thetab)~,
\end{equation}
where $\ab\in\Rd^3$ refer to  the X-ray source location.
For a given X-ray source trajectory $\ab(\lambda)$,  the differentiated backprojection (DBP) to a point $\xb$ on the PI-line  (or chord line) specified by $\lambda^-:=\lambda^-(\xb)$ and $\lambda^+:=\lambda^+(\xb)$ is 
then formally defined as \cite{pack2005cone,zou2004exact,zou2004image}:
\begin{equation}\label{eq:g}
g(\xb)=\int_{\lambda^-}^{\lambda^+}d\lambda~\frac{1}{\|\xb-\ab(\lambda)\|}\left.
\frac{\partial}{\partial\mu}D_f(\ab(\mu),\thetab)\right|_{\mu=\lambda}
\end{equation}
where 
and $1/\|\xb-\ab(\lambda)\|$ is the weighting of distance. 

One of the most important aspects of the DBP formula in \eqref{eq:g} is its relation to the analytic reconstruction methods. More specifically,
the authors in the reference \cite{pack2005cone,zou2004exact,zou2004image} showed that
\begin{eqnarray}\label{eq:K}
f(\xb) = \int_{\Rd^3} d\xb' K(\xb,\xb')g(\xb')
\end{eqnarray}
where 
\begin{eqnarray}
K(\xb,\xb'):= \frac{1}{j2\pi} \int_{\Rd^3}d\nub \sgn(\nub\cdot \eb_\pi(\xb))e^{j2\pi\nub \cdot(\xb-\xb')}
\end{eqnarray}
\begin{eqnarray}
\eb_\pi(\xb):=\frac{\ab(\lambda^+)-\ab(\lambda^-)}{\|\ab(\lambda^+)-\ab(\lambda^-)\|}~.\nonumber
\end{eqnarray}
The authors in \cite{pack2005cone,zou2004exact,zou2004image} further proved that \eqref{eq:K} can be equivalently
represented by
\begin{equation}\label{eq:f_g_hbt2}
f_{v}(u)=-\frac{1}{2\pi}\Hc g_{v}(u) \ .
\end{equation}
where $g_{v}(u)$ and $f_{v}(u)$  are the value of  $g(\xb)$ and $f(\xb)$ on the PI-line index $v$, respectively;
and
$\Hc$ is the Hilbert transform:
\begin{equation}\label{eq:f_g_hbt}
\Hc g(u):=\int\frac{d\eta}{\pi (u-\eta)}g(\eta) \ .
\end{equation}
This  is known as the backprojection filtration (BPF) method, which reconstructs the object $f_{v}(u)$ from the DBP data $g_{v}(u)$ by conducting the Hilbert transform $\Hc$ on each PI  (or chord) line \cite{pack2005cone,zou2004exact,zou2004image}.

\begin{figure}[!hbt]
\centering
\includegraphics[width=7cm]{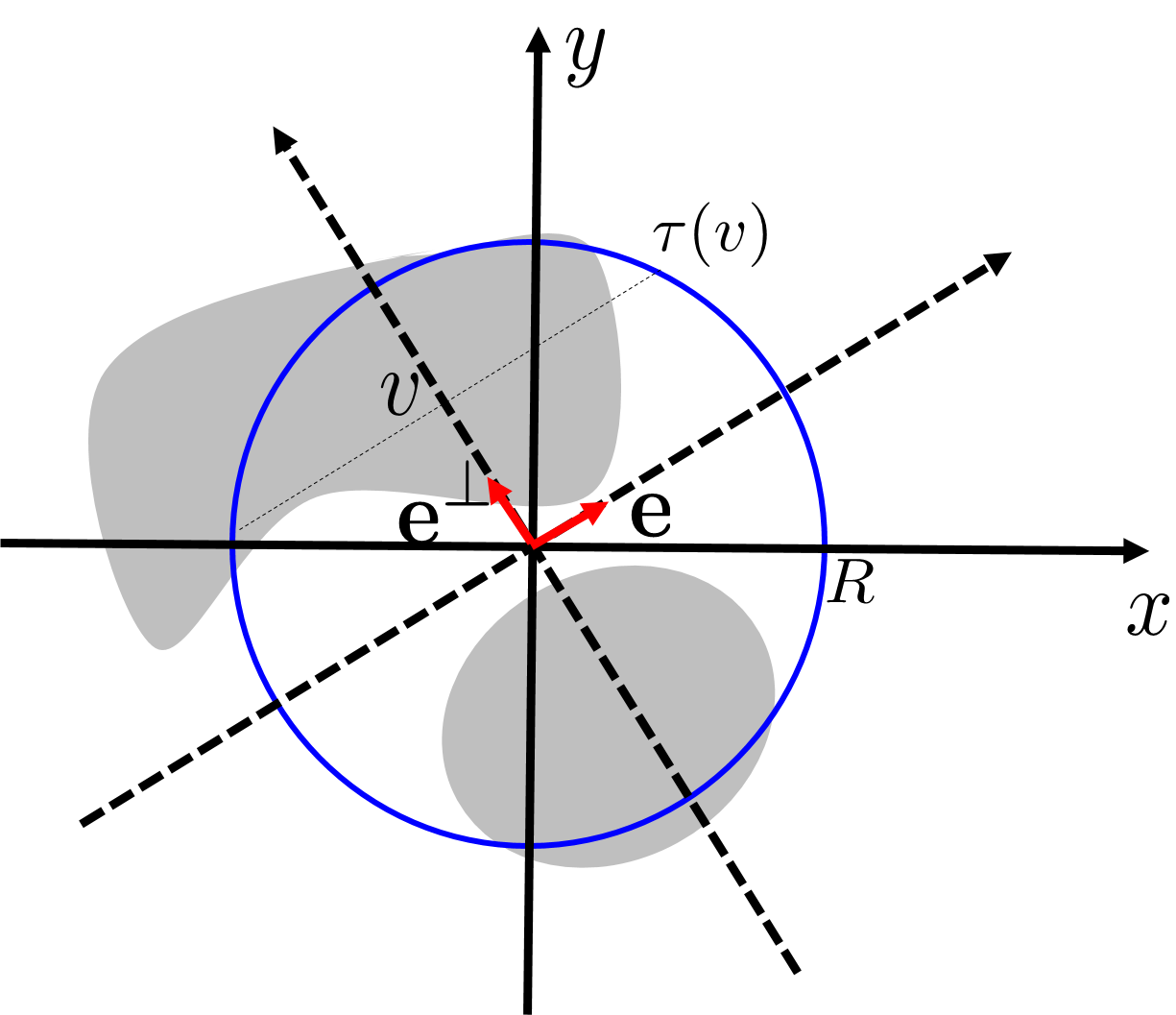}
\caption{\bf\footnotesize A coordinate system for interior tomography.}
\label{fig:coordinate}
\end{figure}

\subsection{Problem Formulation}

The interior tomography then measures the restriction of the Radon measurement $\Rc f$ in the
region $\{(\thetab,s)~:~|s|<R \},$ where $R$ is the radius of the ROI.
In the DBP domain, this is equivalent to  find the unknown $f_v(u), |u|<\tau$ on the chord line indexed by $v$ using
the DBP measurement $g_v(u), |u|<\tau$, where
$\tau:=\tau(v)$ denotes the chord line dependent upon the 1-D restriction of the ROI (see Fig.~\ref{fig:coordinate}).
Formally,  1-D interior tomography problem can be formally stated as
\begin{eqnarray}\label{eq:P}
(P): & \mbox{Find $\Ic_\tau f_v(u)$  such that  $\Ic_\tau g_v(u) =  \Ic_\tau \Tc_\tau  f_v(u)$}
\end{eqnarray}
where $\Ic_\tau$ be the indicator function   between $[-\tau,\tau]$:
$$\Ic_\tau g (u) = \begin{cases} g(u), & |u|\leq \tau \\ 0, &\mbox{otherwise} \end{cases}.  $$
and $\Tc_\tau$ is the finite Hilbert transform  defined as
\begin{eqnarray}\label{eq:Tc}
\Tc_\tau f(u) =  \int_{-\tau}^\tau \frac{f(u')}{\pi(u-u')}du'.
\end{eqnarray}
 In order to obtain 2-D image within the ROI, this problem should be solved for all $|v|<R$.

In  \eqref{eq:P},  note that $g_v$ is the Hilbert transform of the {\em non-truncated} $f_v$ and
one is interested in recovering its {\em truncated} restriction $\Ic_\tau f_v$ from the truncated measurement $\Ic_\tau g_v$,
Accordingly, there are infinitely many $f_v$  that shares the same truncated measurement $\Ic_\tau g_v$. Therefore, the inverse problem
(P) is very ill-posed and cannot be solved by itself.
In fact, the existing interior tomography approaches  \cite{courdurier2008solving,katsevich2012stability,jin2012interior,katsevich2012finite,yu2009compressed,ward2015interior,lee2015interior}  imposed additional constraint such as smoothness or prior knowledge
to $f_v$ to address this ill-posedness.
In the following, we investigate how this ill-posedness can be handled by neural network approaches.

\subsection{Inversion of  Finite Hilbert Transform using Neural Networks}

\begin{figure}[!bt]
\centerline{
\includegraphics[width=10cm]{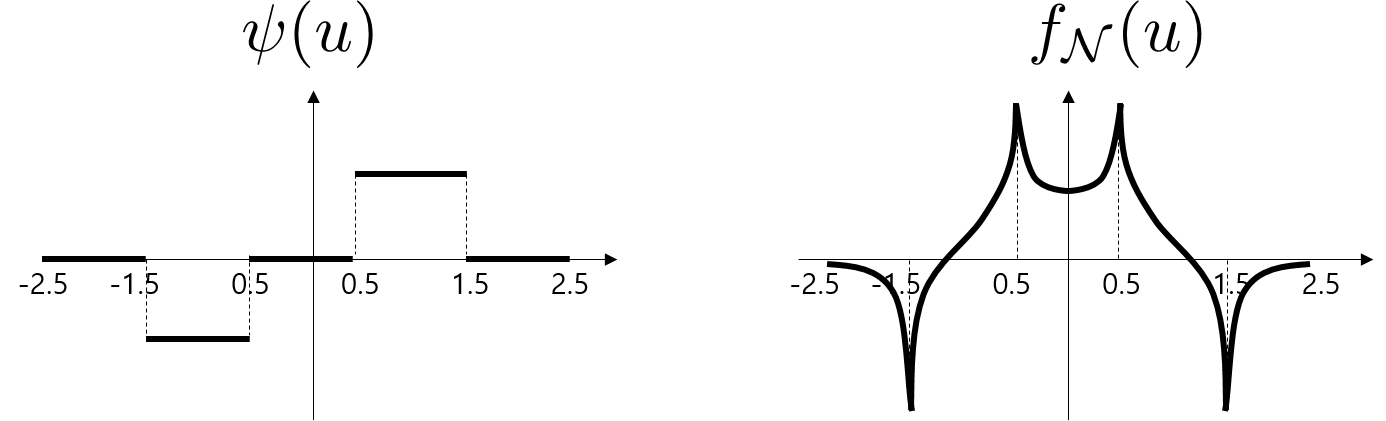} }
\vspace*{-0.5cm}
\centerline{\mbox{(a)}}
\vspace*{0.5cm}
\centerline{
\includegraphics[width=10cm]{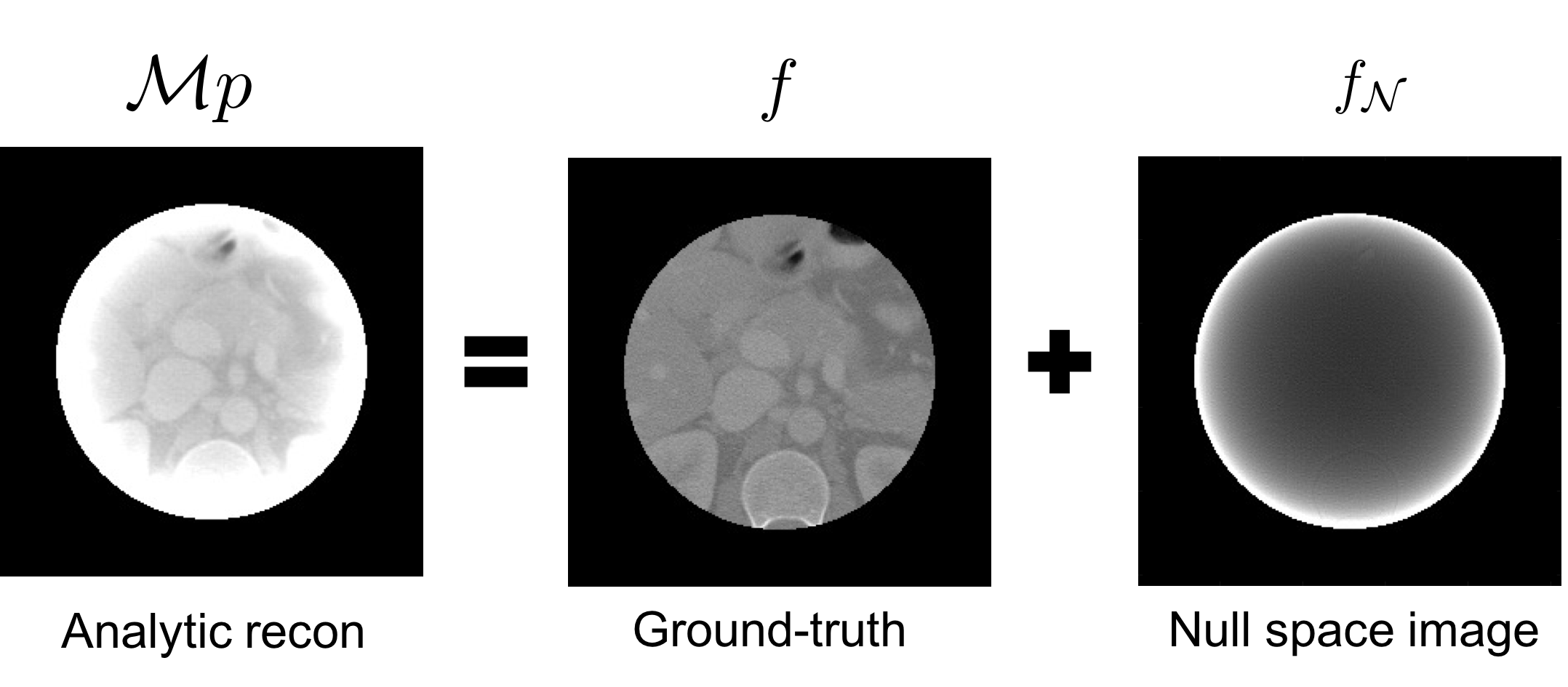} }
\vspace*{-0.3cm}
\centerline{\mbox{(b)}}
\caption{\bf\footnotesize (a) A one-dimensional null space signal, and (b) a 2-D null space image.}
\label{fig:null}
\end{figure}

The main technical difficulty of the interior reconstruction $(P)$ is that
there is a null space in the finite Hilbert transform  \cite{katsevich2012finite,ward2015interior}. 
In particular, there is a non-zero $f_{\Nc}(u)$ such that
\begin{equation*}
\Tc_\tau f_{\Nc}(u)=0,\quad |u|<\tau  \ ,
\end{equation*}
Indeed, $f_{\Nc}(x)$ can be expressed such that
\begin{equation}\label{eq:f_null}
f_{\Nc}(u)=-\frac{1}{\pi}\int_{\Rd\setminus [-\tau,\tau]}\frac{\psi(u')}{u-u'}du'~.
\end{equation}
for any function $\psi(u)$ outside of the ROI.
A typical example of an 1-D null space image $f_{\Nc}$ for a given $\psi$ is illustrated in Fig.~\ref{fig:null}(a) for the case of $\tau=0.5$, 
where the null space signal contains the singularities at $u=\pm 0.5$.
An example of 
2-D image null space image is also shown in Fig.~\ref{fig:null}(b), in which
 the singularities also exists at the ROI boundary.
These are often called as the cupping artifact because they are shaped like a cup with a stronger bias of CT number near the ROI boundary.
The cupping artifacts reduce contrast  and interfere with clinical diagnosis.

\begin{figure}[!t]
\centerline{
	\includegraphics[width=0.7\textwidth]{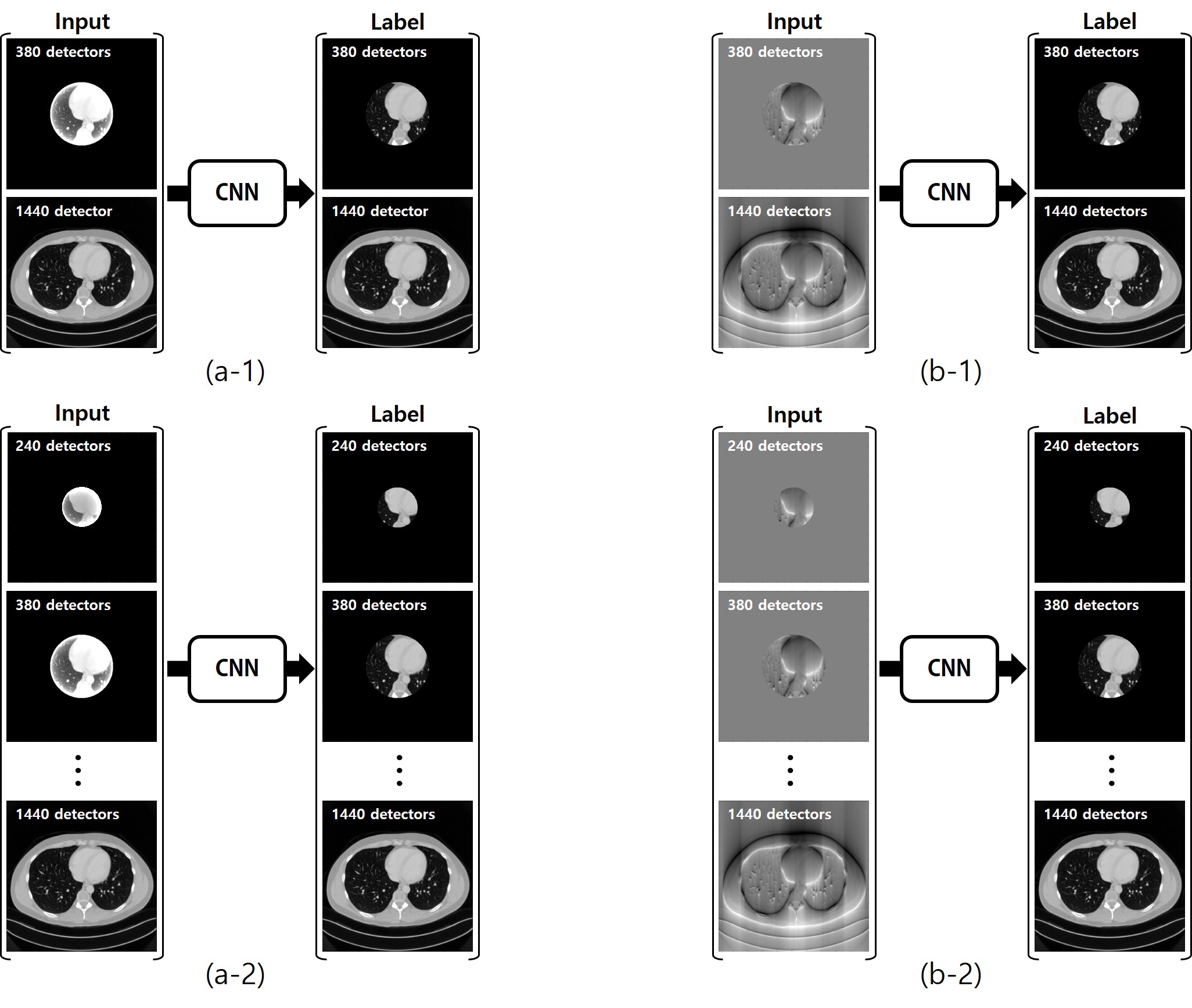} }
\vspace*{-0.5cm}
\caption{\bf\footnotesize Different network architectures for  interior tomography problems. (a) Type I networks trained to
learn the cupping artifacts from the analytic reconstruction, and  (b) type II networks trained to invert  truncated Hilbert transform.  }
	\label{fig:train}
\end{figure}

In the first type of neural network (Type I neural network), which is a direct extension of our preliminary work \cite{han2018roi},   a neural network $\Qc$ is designed such that 
\begin{eqnarray}\label{eq:Q}
\Qc(\fb+\fb_\Nc) = \fb
\end{eqnarray}
where $\fb$ and $\fb_\Nc$ denote the collection of 2-D ground-truth signal and its null space images.
For this network, the null space corrupted input images can be easily obtained by
$$\fb+\fb_\Nc = \Mc \pb,$$
where 
$\Mc$ is 
 the filtered backprojection (FBP)  or convolutional backprojection (CBP) algorithm,
 and $\pb$ denotes the zero-padded truncated projection data.
See Figs.~\ref{fig:train}(a-1)(a-2) for such network.
Then, the neural network training problem  is formulated as
\begin{eqnarray}\label{eq:opt}
\min_{\Qc} \sum_{i=1}^N\|\fb^{(i)} - \Qc \Mc \pb^{(i)} \|^2
\end{eqnarray}
where $\{(\fb^{(i)},\pb^{(i)})\}_{i=1}^N$ denotes the training data set composed of ground-truth image and its truncated projection.
This method is simple to implement, and provides significant gain over the existing iterative methods \cite{yu2009compressed, lee2015interior}.

However, one of the main technical issues of this network architecture is that  the input images are corrupted with
the singularities from the null space images at the ROI boundaries as shown in Fig.~\ref{fig:null}.
Due to the strong intensity at the ROI boundaries,  the network training turn out strongly dependent on ROI size-dependent cupping
artifacts. 
Although this would not be a problem when a specific ROI size is used for all  interior tomography problems, 
in many practical applications such as interventional imaging, cardiac imaging, etc.,  the size of the ROI is mainly dependent on the
subject size and clinical procedures, so there are many demands for flexible ROI sizes during the imaging. 

To design  a neural network that generalizes well for various ROI sizes, let us revisit the finite Hilbert transform \eqref{eq:Tc}. For simplicity, we now assume $\tau=1$ and
$\Tc:=\Tc_1$.
Then, the following formula is well-known as an inversion formula for 
the finite Hilbert transform \cite{king2009hilbert}:
\begin{eqnarray}\label{eq:invH}
\Ic_1 f(u) &=& \frac{c}{\pi\sqrt{1-u^2}}- \frac{1}{\pi\sqrt{1-u^2}}\int_{-1}^1 \frac{\sqrt{1-s^2}\tilde g(s)}{u-s} ds, \quad |u|<1
\end{eqnarray}
where
\begin{eqnarray}\label{eq:tg}
\tilde g:=\Tc_1 f := \frac{1}{\pi}\int_{-1}^1\frac{f(u')}{u-u'}du', & \quad &
c:=  \int_{-1}^1 f(s) ds \ .
\end{eqnarray}
Note that the DBP data from the Hilbert transform can be decomposed as 
\begin{eqnarray*}
g = \Hc f =  \Tc_1 f + \frac{1}{\pi}\int_{|u'|>1}\frac{f(u')}{u-u'}du' = \tilde g - g_\Nc
\end{eqnarray*}
where 
\begin{eqnarray}\label{eq:Nc}
g_\Nc:= -\frac{1}{\pi}\int_{\Rd\setminus [-\tau,\tau]}\frac{f(u')}{u-u'}du'.
\end{eqnarray}
Then,  
we have
\begin{eqnarray}\label{eq:formula}
\Ic_1 f(u) &=&  \frac{\epsilon(u)}{\pi\sqrt{1-u^2}} - \frac{1}{\pi\sqrt{1-u^2}}\int_{-1}^1 \frac{\sqrt{1-s^2} g(s)}{u-s} ds
\end{eqnarray}
where the offset function $\epsilon(u)$ is defined as
\begin{eqnarray}\label{eq:e}
\epsilon(u):=c- \int_{-1}^1 \frac{\sqrt{1-s^2} g_\Nc(s)}{u-s} ds.
\end{eqnarray}
Although the  formula \eqref{eq:formula} with \eqref{eq:e} appear as a desired inversion formula  that  could
be directly used for interior tomography problems,
the main weakness of this formula is that the offset $\epsilon(u)$ is difficult to compute analytically.

To investigate how this problem can be addressed using the second type of neural network (Type II network),  note that the inversion formula can be  
converted to
\begin{eqnarray}
(w_\tau\odot \Ic_\tau f_v)(u)= \epsilon_v(u) - h \ast (w_\tau\odot g)(u), && |v| < \tau
\end{eqnarray}
where  $\tau:=\tau(v)$ denotes the window size for the chord line index $v$, 
$\odot$ denotes the element-wise product, and
$w_\tau(u)$ is the analytic form of weighting given by
\begin{eqnarray}\label{eq:wfactor}
w_\tau(u) = \pi \sqrt{\tau^2-u^2}, \quad |u| \leq \tau, 
\end{eqnarray}
and $h$ is the convolution kernel for Hilbert transform,
and $\epsilon_v(u)$ is the unknown offset function.
Since the analytic weighting $w_\tau$ can be readily calculated once the ROI size is detected from the truncated DBP input,
the required parameters for the reconstruction of $(w_\tau\odot \Ic_\tau f_v)$ is the convolutional kernel $h$ and the offset $\epsilon_v$ for
all chord line index $|v|<\tau$. Then,  after the reconstruction, the weight $w_\tau$ can be removed and the final image $f_v$ can be obtained
for all $|v|<\tau$.

In fact, this algorithmic procedure can be readily learned using a deep neural network.
Specifically, we construct a neural network $\Sc$ such that
$$\fb = \Sc \gb,$$
where $\gb$ denotes the truncated DBP data for all chord lines, and $\fb$ is the 2-D ground-truth image.
The roles of the neural network $\Sc$ are then to estimate the ROI size $R$ (and its restriction $\tau$) from the truncated DBP input
to calculate the  weighting, and to learn
the  convolutional kernel for Hilbert transform as well as the offset $\epsilon_v$ for all $|v|<\tau$.

Such  neural network training problem  can be performed as
\begin{eqnarray}\label{eq:opt2}
\min_{\Sc} \sum_{i=1}^N\|\fb^{(i)} - \Sc \gb^{(i)} \|^2
\end{eqnarray}
where $\{(\fb^{(i)},\gb^{(i)})\}_{i=1}^N$ denotes the training data set composed of ground-truth image  and its 2-D DBP data.
It is desirable if the network can learn to invert Hilbert transform for the full DBP data as well. 
Therefore, both  full DBP data and truncated DBP data as well as  their corresponding ROI ground-truth image should be used as input and label data for the training. 

In contrast to the type I neural network $\Qc$, the type II neural
network $\Sc$ has truncated DBP data as input, which are just ROI images of the full DBP data. So there exists no singularities in the input data.
Later we will show that such a trained neural network has a significant generalization power so that it can be used for any ROI sizes and different
acquisition conditions.

\section{Method}
\label{sec:method}

\subsection{Data Set}

Ten subject data sets from  American Association of Physicists in Medicine (AAPM) Low-Dose CT Grand Challenge were used in this study.
The provided data sets were originally acquired using  helical CTs, and were rebinned  to $360^{\circ}$ angular scan fan-beam CT format. Specifically, the  raw projection data was acquired by a 2D cylindrical detector and a helical conebeam trajectory using a z-flying focal spot \cite{flohr2005image}, so they were first converted into conventional fanbeam projection data using a single-slice rebinning technique \cite{noo1999single}.
The $512 \times 512$ size artifact-free CT images are reconstructed from the rebinned fan-beam CT data using  filtered backprojection (FBP) algorithm.  
For the generation of the training data, sinogram are numerically obtained using forward fan-beam projection operator for our experiments. 
The number of detectors is 1440 elements with pitch of 1 mm. The number of views is 1200. The distance from source to origin (DSO) is 800 mm and the distance from source to detector (DSD) is 1400 mm. The size of images is 512 $\times$ 512.
Out of ten patient, eight patient data were used as training sets, one patient data was used as validation set, and the other patient data was used for test set. 
This corresponds to 3720, 254, and 486 slices for training, validation, and test data, respectively.

Additionally, we used the original raw projection data from AAPM Low-Dose CT Grand Challenge. 
 This data is used only for test phase to validate the generalization performance for real measurement.


Fig. \ref{fig:train}(a) shows a flowchart of the training scheme  for Type I neural network $\Qc$ that learns the artifact-free images from the analytic reconstruction images using the truncated projection data.
In this case, the input image is corrupted with the cupping artifact, whereas the clean data with the same ROI is used as the ground truth. 
In Fig. \ref{fig:train}(a-1), the  truncated input images were generated from 380 detectors, which has a radius of 107.59 mm and about 30\% of the total ROI.
In addition, artifact-free images (1440 detectors) were used simultaneously to improve the generalization performance. 
Fig. \ref{fig:train}(a-2) is another version of Type I neural network. Various ROI images from 240, 380, 600, 1440 detectors were used as the training dataset for network training. This corresponds to the ratio of 19, 30, 46, and 100\%, respectively.

Fig. \ref{fig:train}(b) shows a flowchart of the training scheme for Type II neural networks $\Sc$ that learn the inverse of the finite Hilbert transform.
The truncated DBP data  has no singularities regardless of   the truncation ratio. 
In fact, the truncated DBP data are exactly the same as the full DBP data within the ROI mask. 
We trained two networks. Similar to the training scheme for  Type I neural network, one network was trained with only 380 detectors and full detectors (see Fig. \ref{fig:train}(b-1)), whereas the other network is trained with various ROIs generated by 240, 380, 600, 1440 detectors (see Fig. \ref{fig:train}(b-2)).

Note that our training data was generated from the full-scan synthetic projection data with the detector pitch of 1 mm; but at the test phase, the trained network was also evaluated using the short-scan data, different detector pitch,  image resolution, and/or for real projection data to validate the generalizability of the network.
We also compared our methods with existing iterative methods such as
 total variation penalized reconstruction (TV) \cite{yu2009compressed} and the L-spline based multi-scale regularization method by Lee et al \cite{lee2015interior}. 


For quantitative evaluation,  the peak signal to noise ratio (PSNR) is computed by
\begin{eqnarray}
	PSNR 
		 &=& 20 \cdot \log_{10} \left(\frac{nm\|f^*\|_\infty}{\|\hat f- f^*\|_2}\right) \  ,
\label{eq:psnr}	
\end{eqnarray}
where $\hat{f}$ and $f^*$ indicate the estimated image and ground truth, respectively. $m$ and $n$ are the number of rows and columns. We also used the structural similarity (SSIM) index \cite{wang2004image}, defined as 
\begin{eqnarray}
	SSIM = \dfrac{(2\mu_{\hat f}\mu_{f^*}+c_1)(2\sigma_{\hat f f^*}+c_2)}{(\mu_{\hat f}^2+\mu_{f^*}^2+c_1)(\sigma_{\hat f}^2+\sigma_{f^*}^2+c_2)},
\end{eqnarray}
where $\mu_{\hat f}$ and $\sigma_{\hat f}^2$ are the mean and the variance  of $\hat f$, and $\sigma_{\hat f f^*}$ is the cross-covariance of $\hat f$ and $f^*$, and  $c_1=(k_1L)^2$ and $c_2=(k_2L)^2$ with $L$ denoting a dynamic range of the pixel intensities and $k_1=0.01$ and $k_2=0.03$. 
We also use the normalized mean square error (NMSE).

\subsection{Network Architecture}

\begin{figure}[!t]
\centering
\includegraphics[width=1\textwidth]{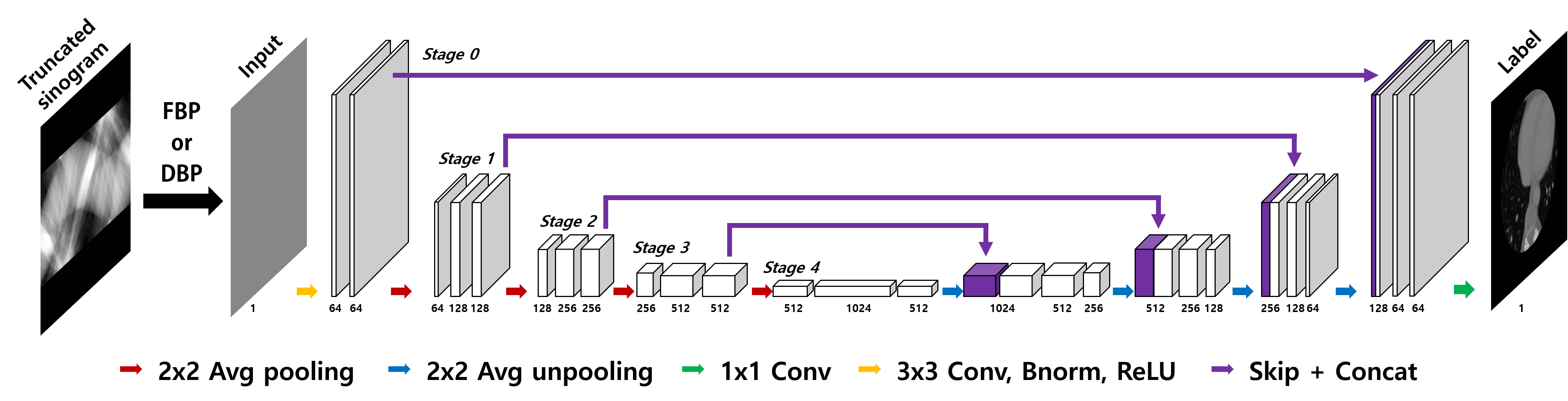} 
\caption{\bf\footnotesize Deep neural network  architecture for the proposed methods.}
\label{fig:network}
\end{figure}

The  network architecture shown in Fig. \ref{fig:network} is used for both Type I and Type II networks, in which  only difference
is from their input images. Type I network uses the FBP images as input, while Type II network uses the DBP data.
The network backbone corresponds to  a modified architecture of U-Net \cite{ronneberger2015u}.
A yellow arrow in Fig. \ref{fig:network} is the basic operator and consists of $3 \times 3$ convolutions followed by a rectified linear unit (ReLU) and batch normalization.
The yellow arrows between the separate blocks at each stage are not shown for simplicity.
A red arrow is a $2 \times 2$ average pooling operator and located between the stages.
Average pooling operator doubles the number of channels and reduces the size of the layers by four.
In addition, a blue arrow  is $2 \times 2$ average unpooling operator, reducing the number of channels by half and increasing the size of the layer by four. 
A violet arrow is the skip and concatenation operator. 
A green arrow is the simple $1 \times 1$ convolution operator generating final reconstruction image. 
The total number of trainable parameters is about 22,000,000.

\begin{figure}[!b]
\centering{
\includegraphics[width=0.6\textwidth]{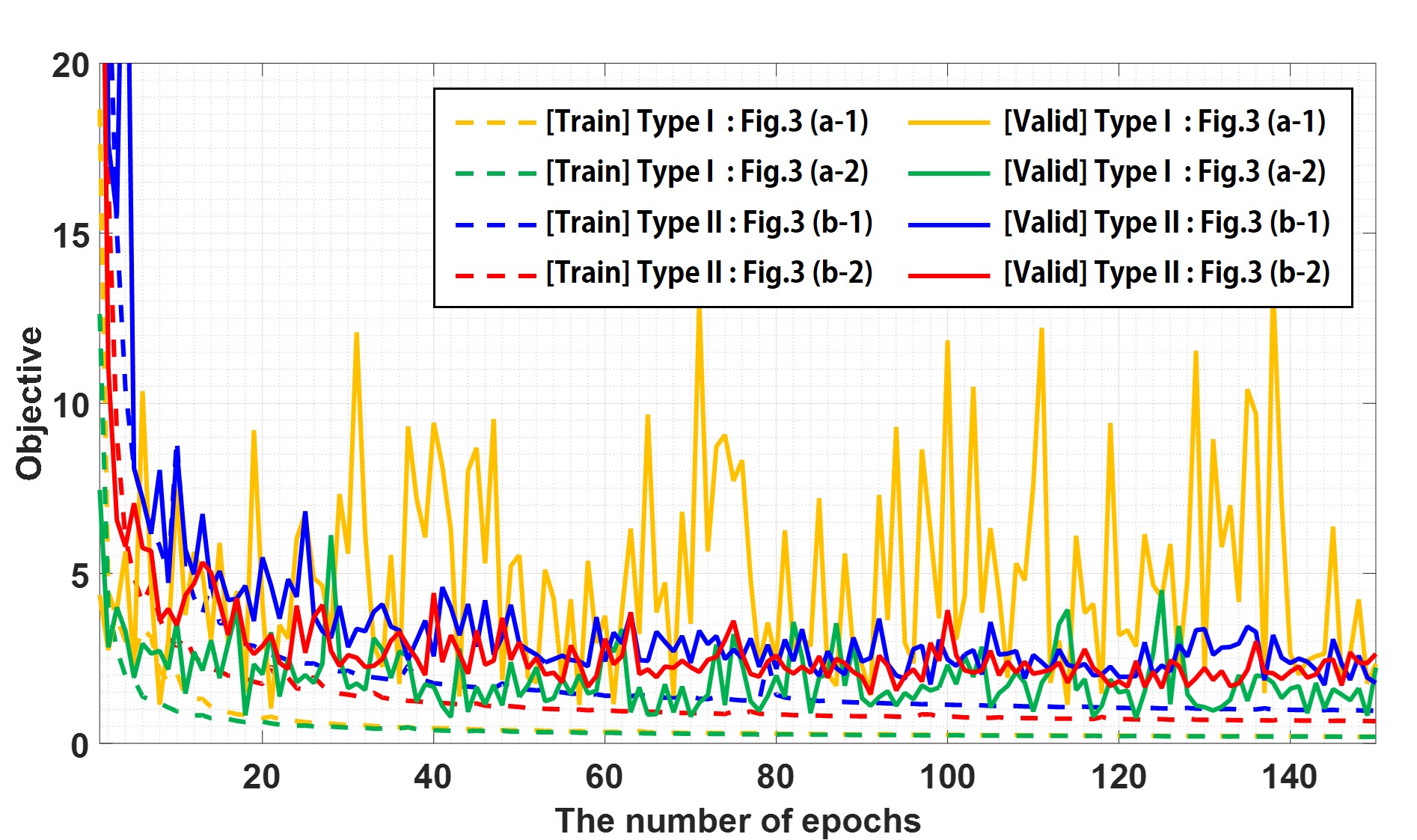} }
\vspace*{-0.5cm}
\caption{\bf\footnotesize Convergence plots for the objective function.  [Train]: training curve. [Valid]: validation curve.}
\label{fig:objective}
\end{figure}

\subsection{Network training}
MatConvNet toolbox (ver.24) \cite{vedaldi2015matconvnet} was used to implement Type I and Type II networks in MATLAB R2015a environment. 
Processing units used in this research are Intel Core i7-7700 (3.60 GHz) central processing unit (CPU) and GTX 1080 Ti graphics processing unit (GPU).
The number of epochs for training networks in Figs.~\ref{fig:train}(a,b)(1) and Figs.~\ref{fig:train}(a,b)(2) were 300 and 150, respectively.
Note that the a-1 and b-1 type networks use the training dataset generated by 380 and 1440 detectors, and the number of training data set  is $3720 \times 2 = 7,440$ slices. However, a-2 and b-2 type networks use $3720 \times 4 = 14,880$ slices because the training dataset were generated by 240, 380, 600, and 1440 detectors. Thus, the number of epochs for a-1 and b-1 type networks should be doubled for fair comparison. 


 Other parameters are same for both types. Stochastic gradient descent (SGD) method was used as an optimizer to train the network. 
The initial learning rate was $10^{-4}$, which gradually dropped to $10^{-5}$ at each epoch. 
The regularization parameter was $10^{-4}$.
For data augmentation, rotated input data of 45$^\circ$, 90$^\circ$, and 135$^\circ$ was added.  This augmentation was necessary
to deal with the short scan acquisition with the different starting angle.
 Therefore, the number of dataset is quadrupled. In addition, the whole data were performed with horizontal and vertical flipping. The mini-batch was used as 4 and the size of input patch is 256 $\times$ 256. Since  convolution operations are spatially invariant, the trained filters were used for the entire 512 x 512 input data at the inferential phase. 
Training time was about 1 week. Fig. \ref{fig:objective} shows the convergence plot for  each network. The dashed and solid lines represent the objective function of the training and validation phases, respectively. Since the training and validation curves converge closely, we concluded that the networks are well-trained and not over-fitted. 

Recall that there are significant differences in the level of cupping artifacts of the FBP images.  For example,
 input images generated using 380 detectors suffer from significant cupping artifacts, whereas the images from 1440 detector geometry are from full detectors so that they do not exhibit any cupping artifact.  Accordingly, Type I network must learn quite distinct distributions simultaneously during the training phase, which makes the objective function of Type I network fluctuates as shown  in Fig. \ref{fig:objective}.
On the other hand, the artifact patterns for the DBP input images from various detectors are not too different so that the training with both input images exhibits the stable convergence behavior.  



\section{Experimental Results}
\label{sec:result}

Due to the null space image that has singularity in the ROI boundary, Type I
network training with analytic reconstruction is quite dependent upon  the input ROI size.
Thus, we conjectured that Type I network with a specific ROI 
may not generalize well for other ROI sizes.

\begin{figure}[!b]
\centering {
\includegraphics[width=0.7\textwidth]{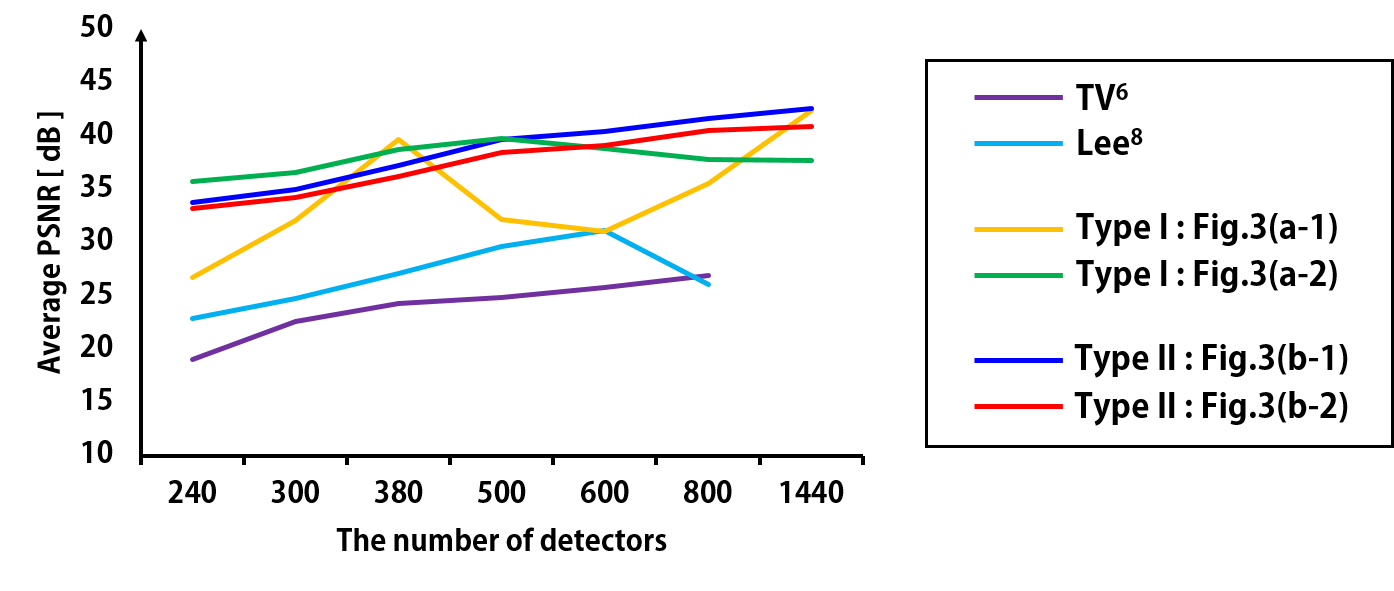} }
\vspace*{-0.7cm}
\caption{\bf\footnotesize Average PSNR  with respect to various ROI sizes determined by the number of detectors.}
\label{fig:result_grp}
\end{figure}

\begin{table}[]
\caption{\bf\footnotesize Quantitative comparison with respect to various ROI sizes determined by the number of detectors.}
\vspace*{-0.5cm}
\label{tbl:psnr_ssim}
\begin{center}
\begin{tabular}{cc|cccccc}
\hline
\multicolumn{2}{c|}{\multirow{2}{*}{PSNR {[}dB{]}}}	& \multirow{2}{*}{{TV} \cite{yu2009compressed}}	& \multicolumn{1}{c}{\multirow{2}{*}{Lee \cite{lee2015interior}}} & \multicolumn{2}{c}{Type I}	& \multicolumn{2}{c}{Type 2}	 \\
\multicolumn{2}{c|}{}	&  	& \multicolumn{1}{c}{} 	& Fig.~\ref{fig:train}(a-1) & Fig.~\ref{fig:train}(a-2)	& Fig.~\ref{fig:train}(b-1) 	& Fig.~\ref{fig:train}(b-2)          \\ \hline\hline
\multirow{7}{*}{\rotatebox[origin=c]{90}{\# of detectors}}
& 240  	& 19.0360   	& 22.8487     	& 26.7065   	& 35.7028  		& 33.7777 		& 33.1967  	\\
& 300  	& 22.6322    	& 24.7433    	& 32.0472     	& 36.5903     	& 35.0179      	& 34.1981  	\\
& 380 	& 24.2809    	& 27.0543    	& 39.6532      	& 38.7082     	& 37.1791      	& 36.1787   \\
& 500  	& 24.8514 		& 29.6371     	& 32.1916 		& 39.7999 		& 39.6883 		& 38.4653 	\\
& 600 	& 25.7784    	& 31.1304      	& 31.0151     	& 38.8380     	& 40.3747     	& 39.1327 	\\
& 800 	& 26.8914 		& 26.0697    	& 35.5274 		& 37.8209 		& 41.6653 		& 40.4745 	\\
& 1440 	& -       		& -            	& 42.3486 		& 37.7013 		& 42.5565 		& 40.9145 	\\ \hline\hline
\multicolumn{2}{c|}{\multirow{2}{*}{SSIM}}         & \multirow{2}{*}{{TV} \cite{yu2009compressed}}    & \multicolumn{1}{c}{\multirow{2}{*}{Lee \cite{lee2015interior}}} & \multicolumn{2}{c}{Type I}                               & \multicolumn{2}{c}{Type 2}                                \\
\multicolumn{2}{c|}{}	&  	& \multicolumn{1}{c}{} 	& Fig.~\ref{fig:train}(a-1) & Fig.~\ref{fig:train}(a-2)	& Fig.~\ref{fig:train}(b-1) 	& Fig.~\ref{fig:train}(b-2)          \\ \hline\hline
\multirow{7}{*}{\rotatebox[origin=c]{90}{\# of detectors}}
& 240  	& 0.7895   		& 0.9445     	& 0.9517   		& 0.9821  		& 0.9727 		& 0.9737  	\\
& 300  	& 0.8504    	& 0.9532    	& 0.9737     	& 0.9849     	& 0.9756      	& 0.9760  	\\
& 380 	& 0.8797    	& 0.9588    	& 0.9889      	& 0.9871     	& 0.9790      	& 0.9792   	\\
& 500  	& 0.8955 		& 0.9624     	& 0.9723 		& 0.9788 		& 0.9817 		& 0.9806 	\\
& 600 	& 0.9019    	& 0.9685      	& 0.9447     	& 0.9703     	& 0.9819     	& 0.9804 	\\
& 800 	& 0.9077 		& 0.9077    	& 0.9207 		& 0.9604 		& 0.9776 		& 0.9771 	\\
& 1440 	& -       		& -            	& 0.9484 		& 0.8296 		& 0.9652 		& 0.9605 	\\ \hline\hline
\end{tabular}
\end{center}
\end{table}

\begin{figure}[!t]
\centering {
\includegraphics[width=0.7\textwidth]{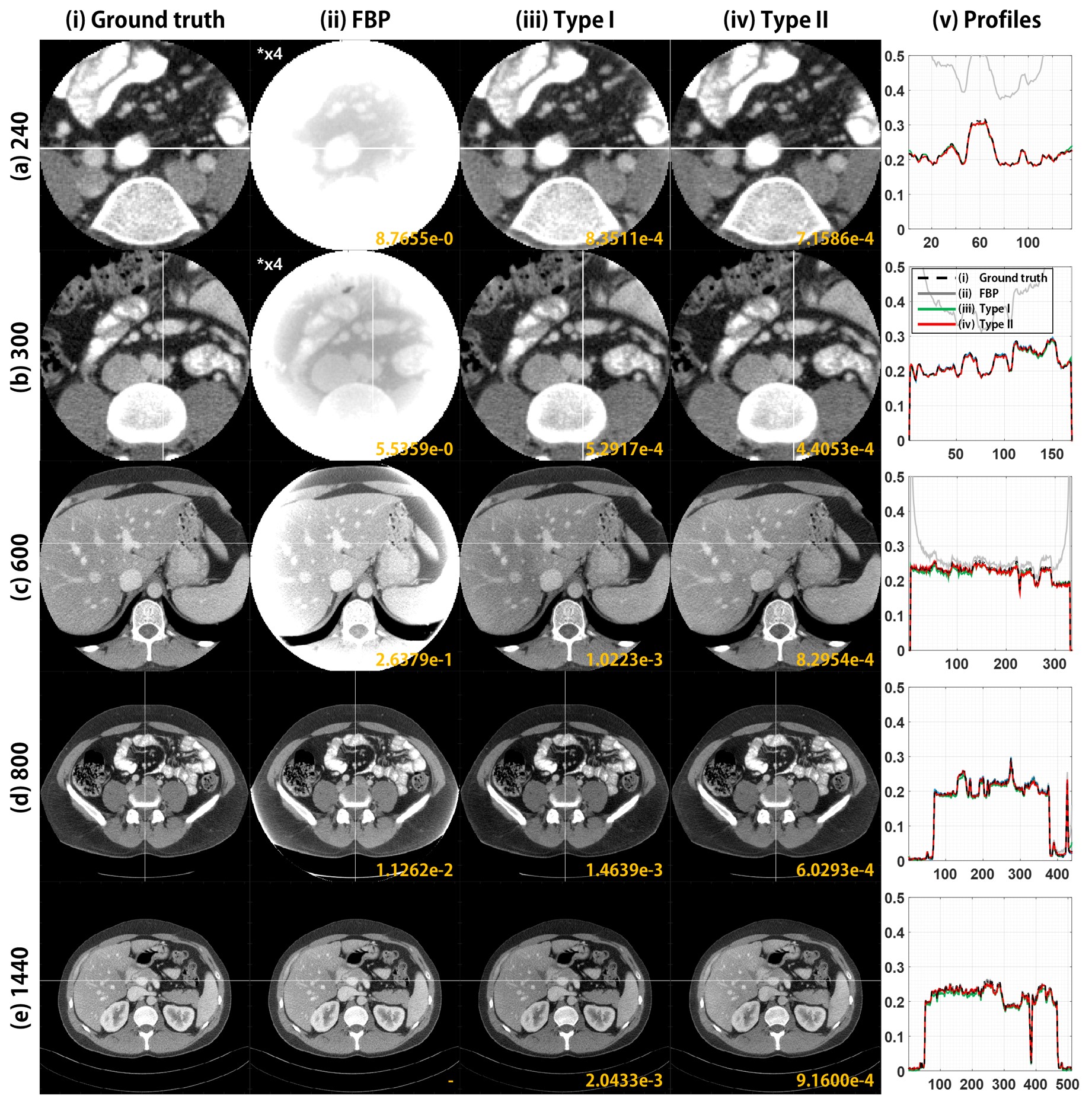} }
\vspace*{-0.5cm}
\caption{\bf\footnotesize Row direction: (i) ground-truth, and (ii) FBP images.  Reconstruction images by (iii) Type I network in Fig.~\ref{fig:train}(a-2), and (iv) Type II neural network in Fig.~\ref{fig:train}(b-2).   (v) shows the profiles indicated with the white line on the results. Column direction:  (a-d) interior images from 240, 300, 600, and 800 detectors, and (e) full 1440 detector image. The NMSE values are written at the corner. A window range is (-150, 300)[HU]. 
The image marked $^*$x4 indicates that the window level is magnified four times.}
\label{fig:result_axial_dct}
\end{figure}

To confirm the performance degradation of Type I network with respect to varying  ROI sizes, the trained network
was applied to the test data for different ROIs from 240 to 1440 detectors.
The average PSNR and SSIM values are described in Fig. \ref{fig:result_grp} and Table \ref{tbl:psnr_ssim}. 
Recall that  Fig. \ref{fig:train}(a-1) were trained only for 380 and 1440 detectors, so it performed best
for the case of the 380 and full detectors as shown in Fig. \ref{fig:result_grp} and Table \ref{tbl:psnr_ssim}. 
However, the performance quickly degrades for different ROI sizes.
While the network in Fig. \ref{fig:train}(a-2)  showed
better performance across  various  ROIs, the performance degradation at 1440 detectors clearly reflects that
the network does not have enough generalization power.

On the other hand, although Type II network
in Fig. \ref{fig:train}(b-1) is also trained only with 380 and 1440 detectors similar to Fig. \ref{fig:train}(a-1), 
it produce robust performance over across various ROI, resulting in  about 7 dB improvement over the conventional iterative methods such as TV \cite{yu2009compressed} and  Lee method \cite{lee2015interior}.
By training with 240, 380, 600, 1440 detectors, the performance of Type II network (i.e. Fig. \ref{fig:train}(b-2))
is about the same as shown in Fig. \ref{fig:result_grp} and Table \ref{tbl:psnr_ssim}.
The  SSIM values from two Type II networks in Figs. \ref{fig:train}(b-1)(b-2)  are  beyond the 0.97 as shown in Table \ref{tbl:psnr_ssim}, implying that both methods can be used for clinical applications, whereas the SSIM value for Type I network in Fig. \ref{fig:train}(a-1) quickly degrades.

In Fig.~\ref{fig:result_axial_dct},  we provide reconstruction results using the network 
 Fig. \ref{fig:train}(a-2)  and Fig. \ref{fig:train}(b-2) for various ROI sizes.
Both networks results in non-distinguishable differences in their reconstruction profiles 
across various ROI sizes.

%


%

\begin{figure}[t]
\centering
\subfigure[~Axial views of reconstruction results from truncated images by 500 detectors.] {
	\includegraphics[width=0.8\textwidth]{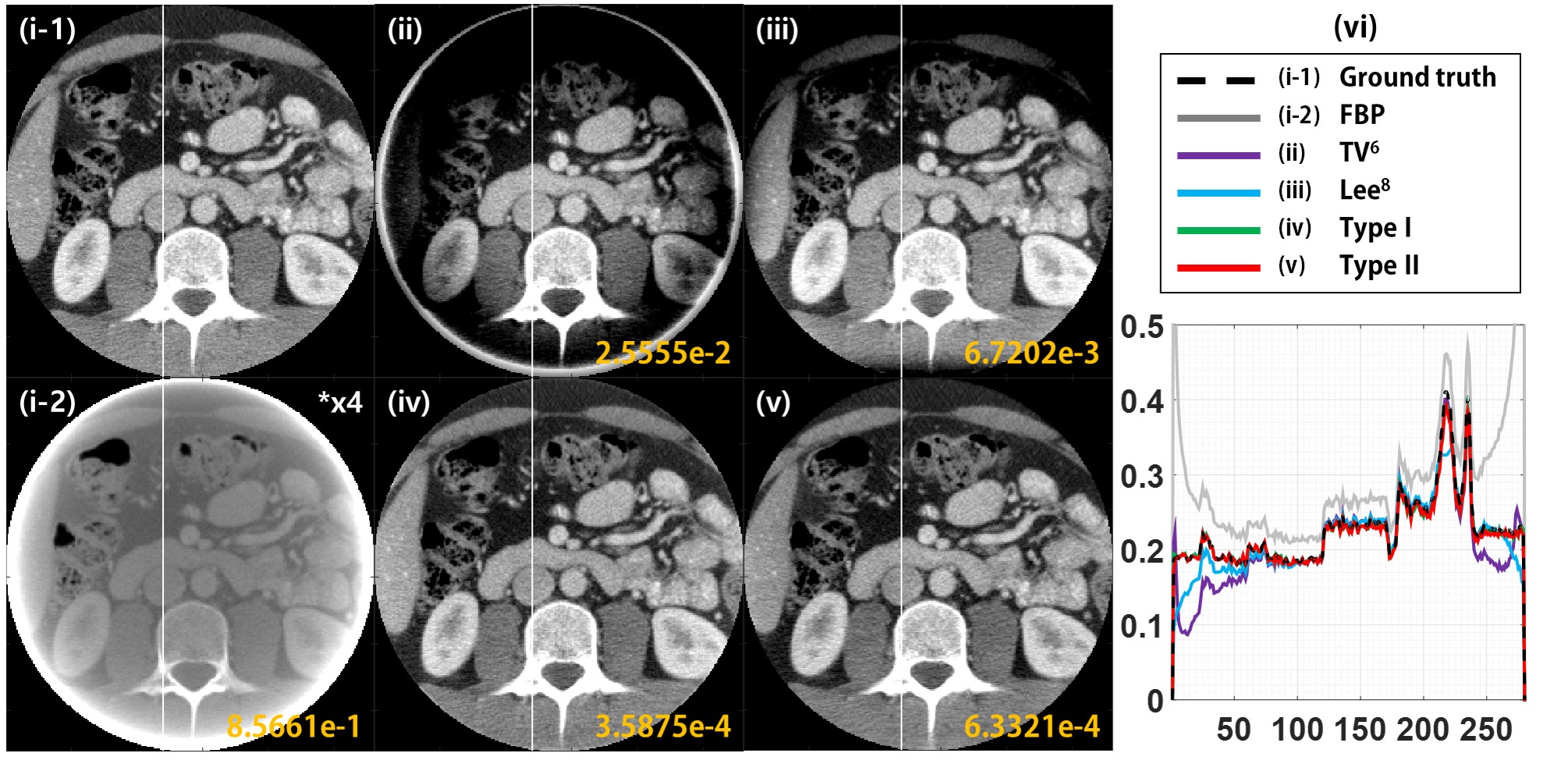} 
	\label{fig:result_500_axial}
}
\centering
\subfigure[~Coronal views of reconstruction results from truncated images by 500 detectors.] {
	\includegraphics[width=0.8\textwidth]{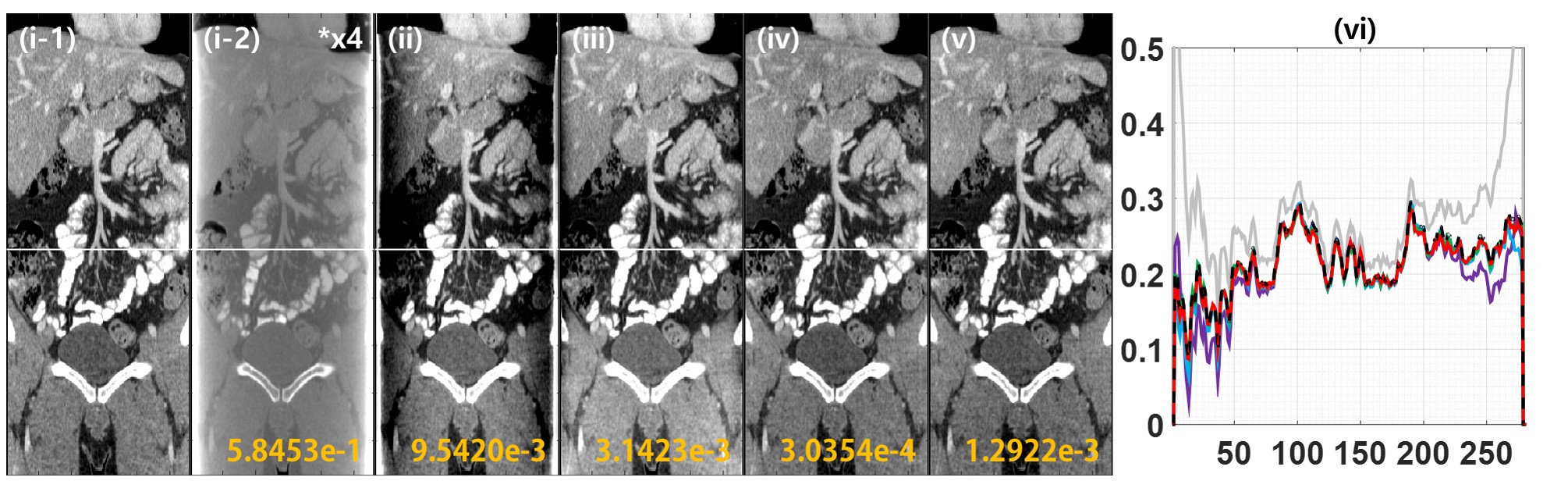} 
	\label{fig:result_500_coronal}
}
\caption{\bf\footnotesize  (i-1) Ground-truth, and (i-2) FBP image. Reconstruction results  by (ii) TV \cite{yu2009compressed}, (iii) Lee method \cite{lee2015interior}, (iv) Type I network, and (v) Type II network from truncated ROI
corresponding to 500 detectors.  (vi)  Reconstruction profiles indicated by the white line on the results. The NMSE values are written at the corner. A window range is (-150, 300)[HU]. The image marked $^*$x4 indicates that the window level is magnified four times.}
\label{fig:result_500}
\end{figure}

\begin{figure}[!t]
\centering {
\subfigure[ ] {
\includegraphics[width=0.7\textwidth]{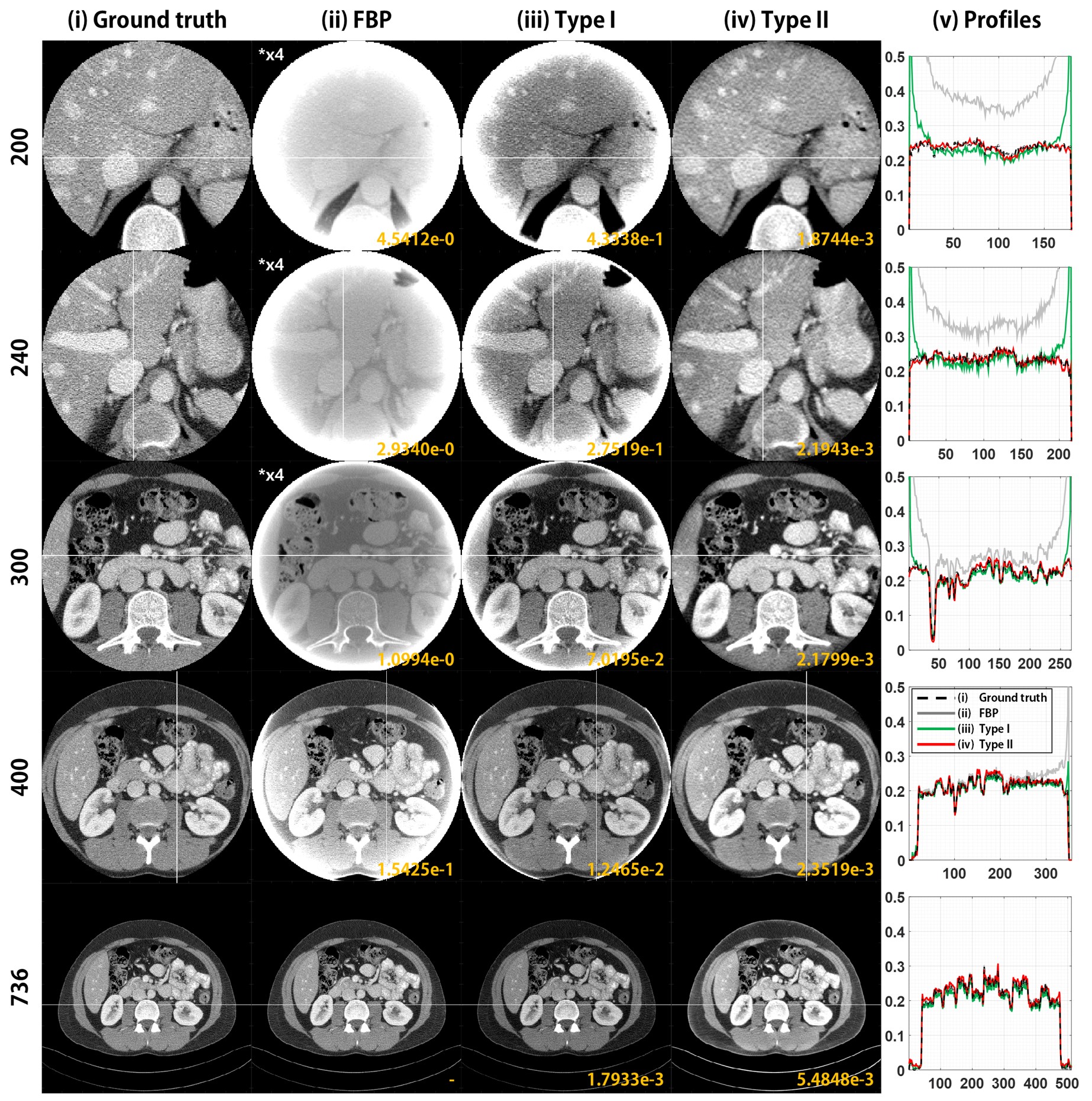} }
\vspace*{-0.5cm}
\label{fig:result_axial_dct_realdata}
}
\centering {
\subfigure[ ] {
\includegraphics[width=0.7\textwidth]{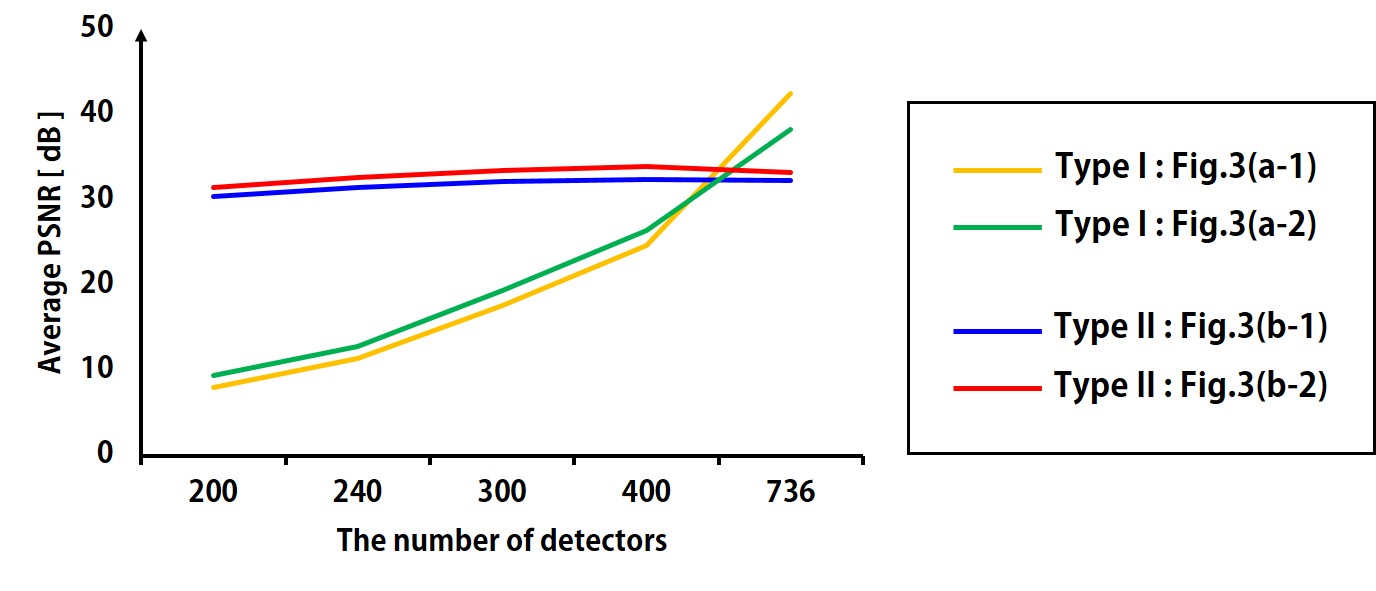} }
\vspace*{-0.5cm}
\label{fig:grp_dct_real_data}
}
\caption{\bf\footnotesize (a) Reconstruction images from real projection measurement, and
 (b) average PSNR values change with respect to the number of detectors. The NMSE values are written at the corner.  A window range is (-150, 300)[HU]. The image marked $^*$x4 indicates that the window level is magnified four times.}
\label{fig:result_axial_dct_real}
\end{figure}


{Fig. \ref{fig:result_500}(i-v) shows the comparison with various methods.
 The graphs in Fig. \ref{fig:result_500}(vi) are the profiles along the white line on the each result.  
 Figs. \ref{fig:result_500}(a)(b) shows the reconstruction results from axial and coronal directions, respectively.
The results clearly show that Type I and Type II networks trained  with diverse ROI data sets (i.e. Fig. \ref{fig:train}(a-2)  and Fig. \ref{fig:train}(b-2)) removed the cupping artifact, and preserved sophisticated structures and textures of the underlying images. The reconstruction profiles also confirmed that the detailed structures are very well preserved by both networks. However, TV method has
residual artifacts at the ROI boundaries, and the Lee method showed a drop in intensity at the ROI boundary.  }

Table \ref{tbl:time} shows the computation time. The GPU implementations of the proposed networks took about 0.05 sec/slice; and on the CPU implementation, they took 4 sec/slice. However, the TV approach took about 11.5 sec/slice on the GPU and the Lee method implemented
on CPU took about 3 $\sim$ 9 sec/slice.  Because the Lee method  is based on a one-dimensional operation, it is faster than TV on the GPU, even though the Lee approach is implemented on the CPU.
 The proposed method in the GPU environment is about 60 times faster than other methods. In addition, the proposed method is 1.5 times faster on the average CPU environment. This confirms that the proposed method, regardless of the ROI sizes, shows very fast reconstruction times and provides remarkably improved image qualities compared to conventional methods.

\begin{table}[t!] 
\caption{\bf\footnotesize Computation times with respect to various detector sizes.}
\vspace*{-0.5cm}
\label{tbl:time}
\begin{center}
\begin{tabular}{cc|cccc}
\hline
\multicolumn{2}{c|}{\multirow{2}{*}{Time [sec/slice]}} & {TV \cite{yu2009compressed}} & Lee \cite{lee2015interior} & \multicolumn{2}{c}{Proposed} \\
 & & GPU & CPU 							  & CPU & GPU \\
\hline\hline
\multirow{4}{*}{\rotatebox[origin=c]{90}{\# of detectors}}
& 240 & 11.6	& 3.3	& \multirow{4}{*}{4.0} & \multirow{4}{*}{0.05} \\
& 380 & 11.7	& 5.1	&  &  \\
& 600 & 11.9	& 9.3	&  &  \\
& 1440& - 		& - 	&  &  \\
\hline
\end{tabular}
\end{center}
\end{table}

To demonstrate the clinical feasibility of the trained networks, the networks trained using synthetic projection data
was used for  real projection data. As for real sinogram measurement, we used  raw projection data from AAPM Low-dose Grand Challenge. Fig. \ref{fig:result_axial_dct_real} shows  the reconstruction images and PSNR curves
  for various ROI sizes using  200, 240, 300, 400, and 736  detectors. 
  Note that the characteristic of cupping artifact is different from the training dataset, since the acquisition parameters for real CT projection data
  are different from the training data.  For example, our training dataset was generated assuming equal-spaced fan-beam, whereas the real AAPM projection data was acquired
 from equal-angular fan-beam with detector offsets. 
The results  in Fig. \ref{fig:result_axial_dct_real}(a) showed that  Type I networks produced artifact,
whereas near artifacts-free images were obtained using Type II network. 
Among the two Type II networks, the network in  Figs. \ref{fig:train}(b-2) that are trained with diverse ROI sizes produced consistent improvement.
This clearly exhibit the generalizability of Type II network.

In clinical CT imaging workflow, images may be reconstructed with any arbitrary pixel size, and even with different
starting angle when used for
short scan acquisition. In this case, the artifact patterns and/or their orientation  depends 
 the selection of detector pitch size and angular range. 
 To investigate the effects caused by system parameters such as pixel sizes, detector pitches, the number of views, and  the starting angles for short-scan, we test the trained networks using data set
 from  various acquisition scenario that are different from training phase.

\begin{figure}[!h]
\centering
\subfigure[~Reconstruction images from various pixel size when 500 detector.] {
\includegraphics[width=0.8\textwidth]{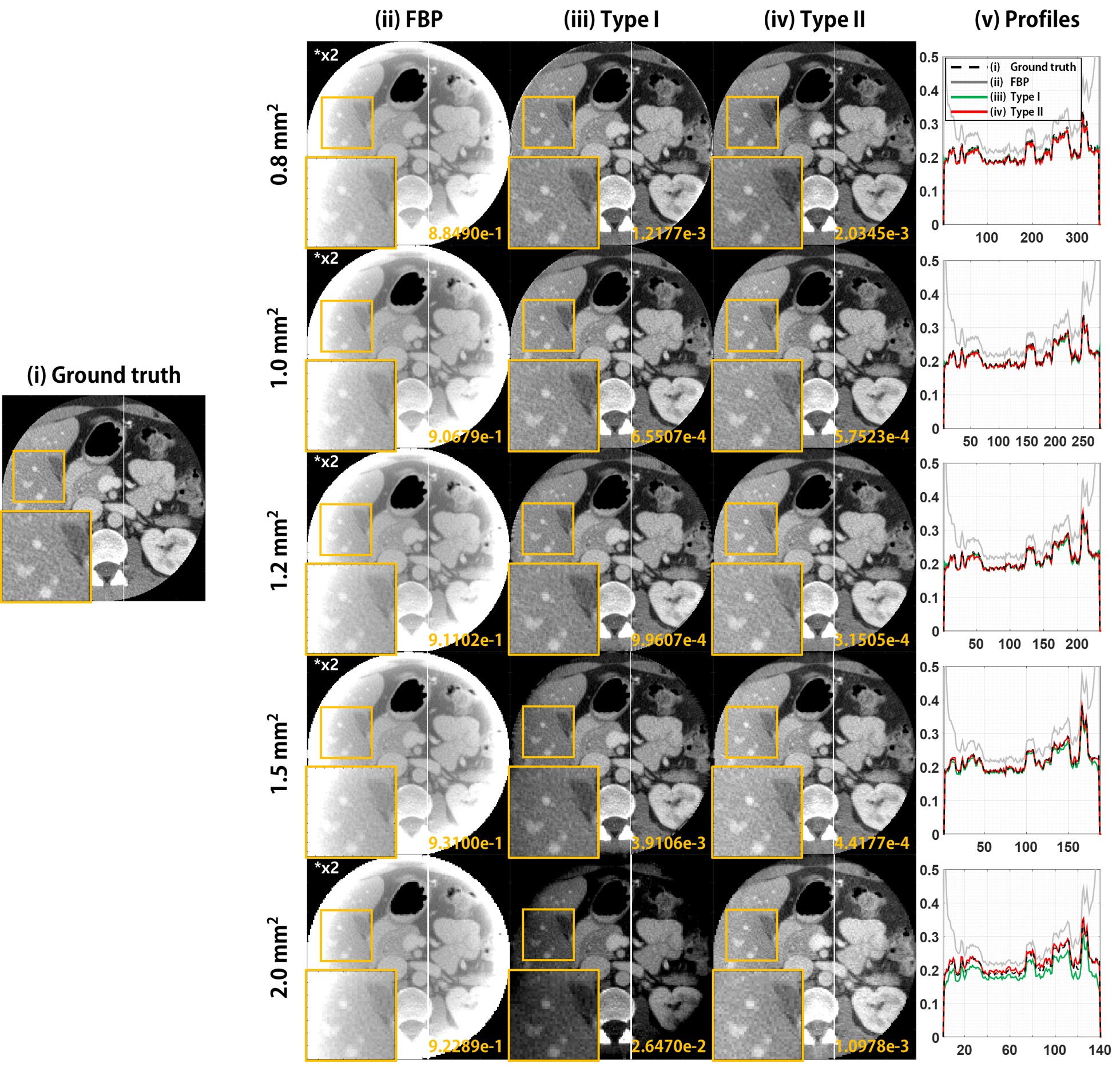} 
}

\centering
\subfigure[~\# of detectors = 380 ] {
	\includegraphics[width=0.3\textwidth]{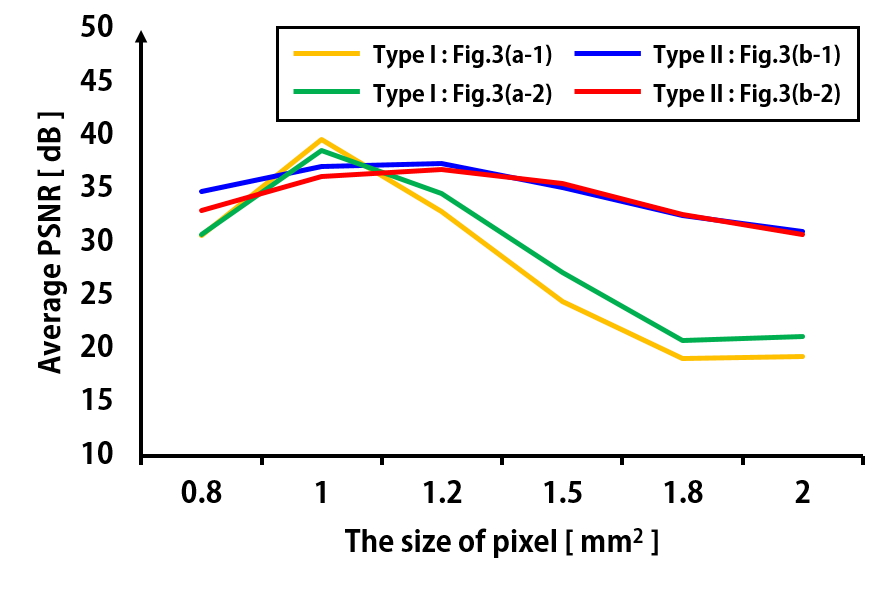} 
	\label{fig:grp_pixel_dct380-0}
}
\centering
\subfigure[~\# of detectors = 500] {
	\includegraphics[width=0.3\textwidth]{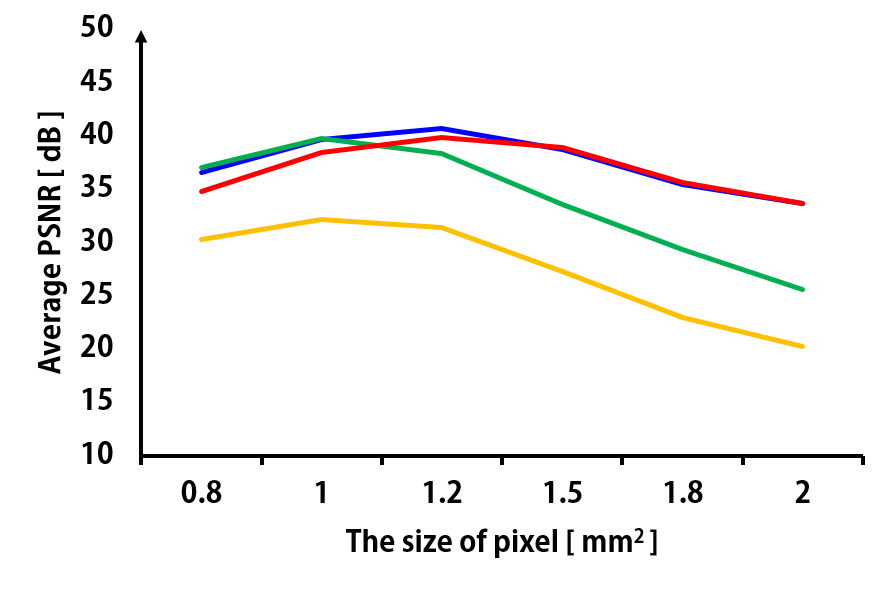} 
	\label{fig:grp_pixel_dct500-0}
}
\centering
\subfigure[~\# of detectors = 600] {
	\includegraphics[width=0.3\textwidth]{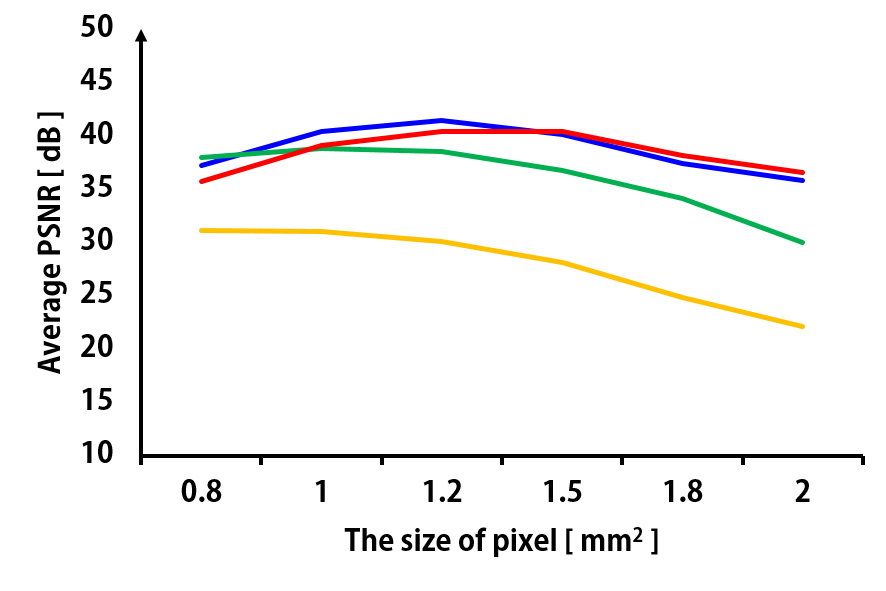} 
	\label{fig:grp_pixel_dct600}
}

\caption{\bf\footnotesize (a) Reconstruction results  under various pixel resolutions, when
 the ROI size corresponds to  500 detectors. (b-d) Average PSNR value with respect to pixel resolution. The number of detectors is (b) 380, (c) 500, and (d) 600, respectively. The NMSE values are written at the corner. A window range is (-150, 300)[HU]. The image marked $^*$x4 indicates that the window level is magnified two times. }
\label{fig:result_axial_pixel_dct}
\end{figure}

{ Fig. \ref{fig:result_axial_pixel_dct} describes the effect of image pixel sizes. Fig. \ref{fig:result_axial_pixel_dct}(a) shows the reconstruction images from the various pixel sizes when the number of detector is 500. When  the pixel size increases,  Type I network tends to under-estimate.  On the other hand,
Type II network provides near perfect reconstruction.
These phenomenon was consistently observed when we change the ROI size.
Figs. \ref{fig:result_axial_pixel_dct}(b)-(d) showed
that
 Type I network performed best for the matched pixel size, but the performance quickly degrades when the reconstruction pixel
 size differs from the training data. 
 On the other hand, Type II network was robust for different pixel sizes.
}

\begin{figure}[!h]
\centering
\subfigure[~Reconstruction images for various detector pitch for ROI = 215.17 mm.] {
\includegraphics[width=0.8\textwidth]{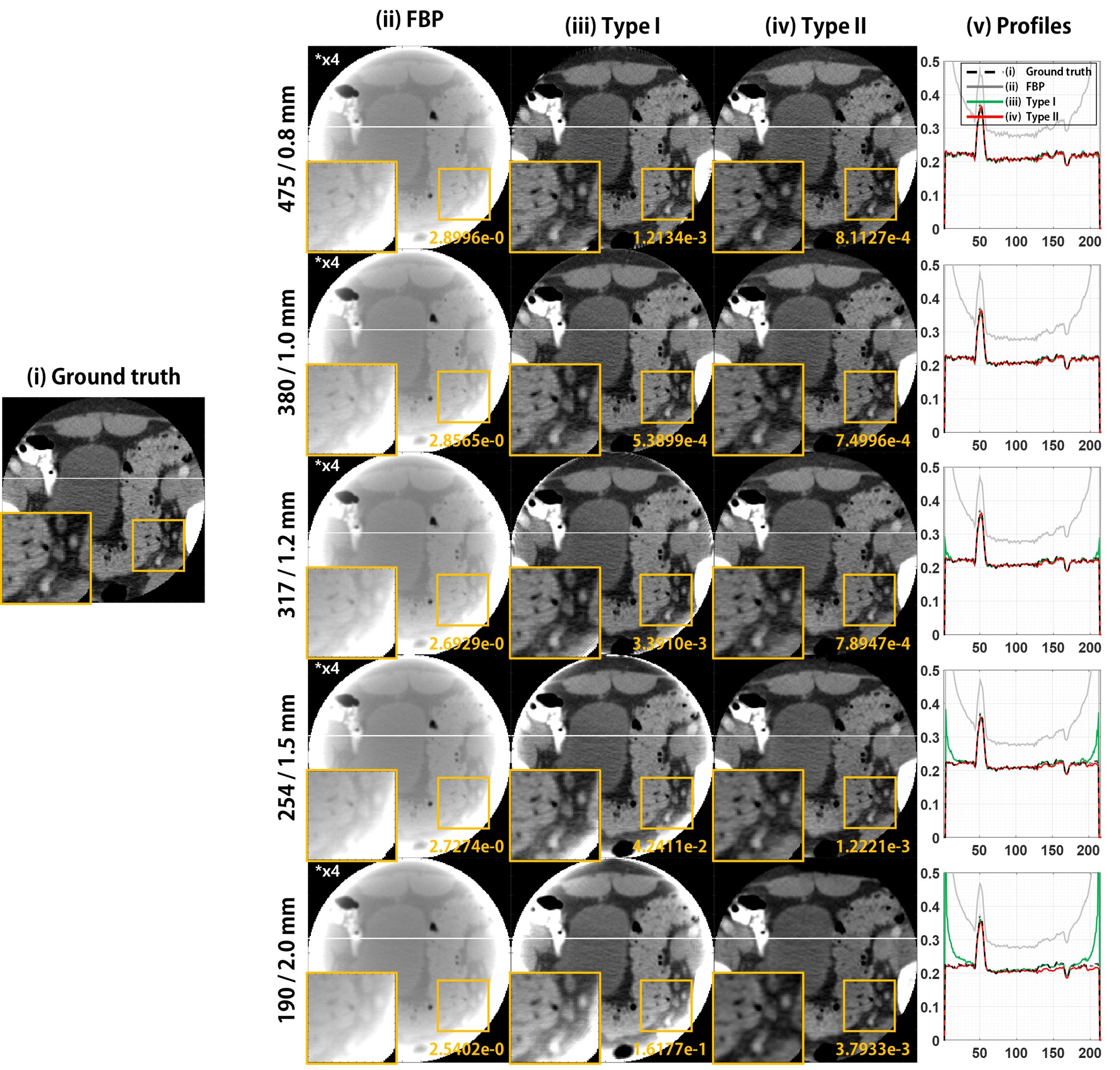} 
}
\centering
\subfigure[~ROI = 170.45 mm] {
	\includegraphics[width=0.3\textwidth]{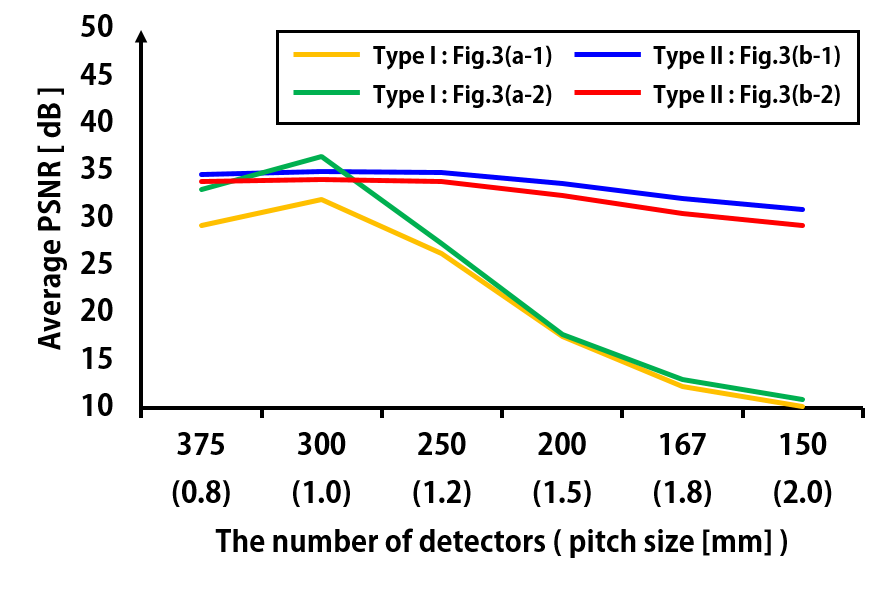} 
	\label{fig:grp_pixel_dct380}
}
\centering
\subfigure[~ROI = 215.17 mm] {
	\includegraphics[width=0.3\textwidth]{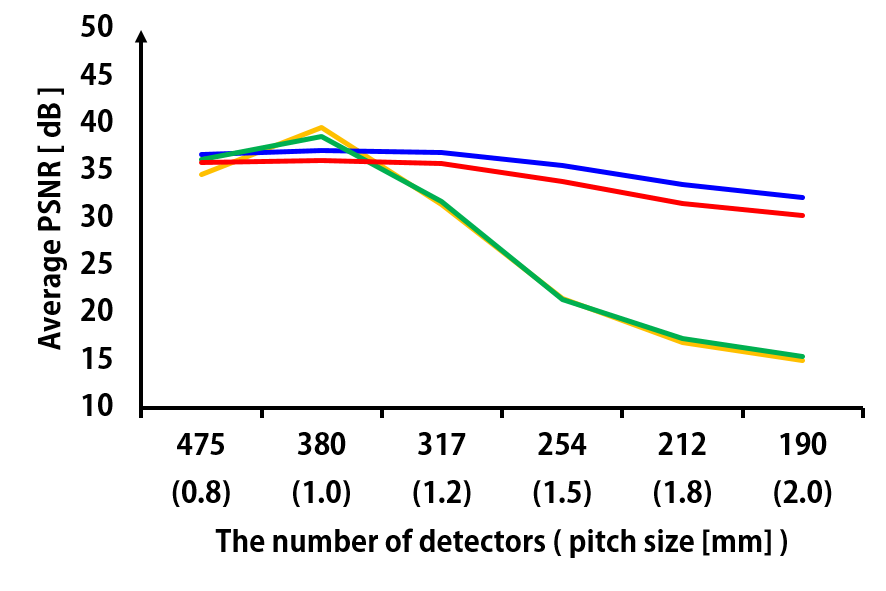} 
	\label{fig:grp_pixel_dct500}
}
\centering
\subfigure[~ROI = 335.25 mm] {
	\includegraphics[width=0.3\textwidth]{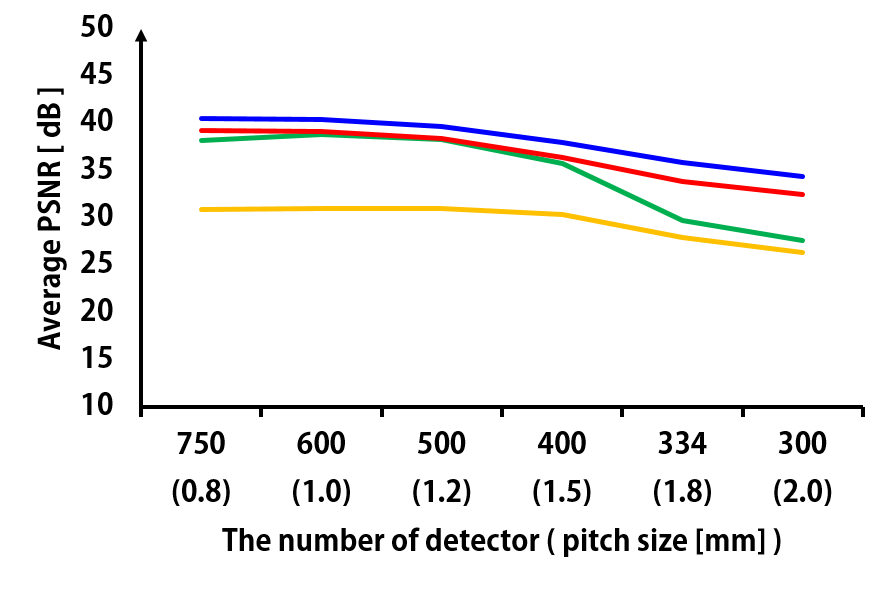} 
	\label{fig:result_axial_fixed_dct}
}

\caption{\bf\footnotesize (a) Reconstruction images for various detector pitch for ROI = 215.17 mm.  (b-d) Average PSNR values with respect to various detector pitch.
The ROI size is (a) 170.45 mm, (b) 215.17 mm, and (c) 335.25 mm, respectively. The NMSE values are written at the corner. A window range is (-150, 300)[HU]. The image marked $^*$x4 indicates that the window level is magnified four times.}
\label{fig:result_axial_pitch_dct}
\end{figure}

Fig. \ref{fig:result_axial_pitch_dct} illustrates the influence of detector pitch  for 
the same ROI size.
While Type I network tends to over-estimate the image value as the pitch size increases, this overestimation
was not seen in results from  Type II network.
Similar phenomenon was observed by changing ROI size and Type II network still exhibits robust performance across various pitch size as shown in Fig. \ref{fig:result_axial_pitch_dct}(b-d).

\begin{figure}[!h]
\centering
\subfigure[~Reconstruction images from various projection view using 240 detectors.] {
\includegraphics[width=0.8\textwidth]{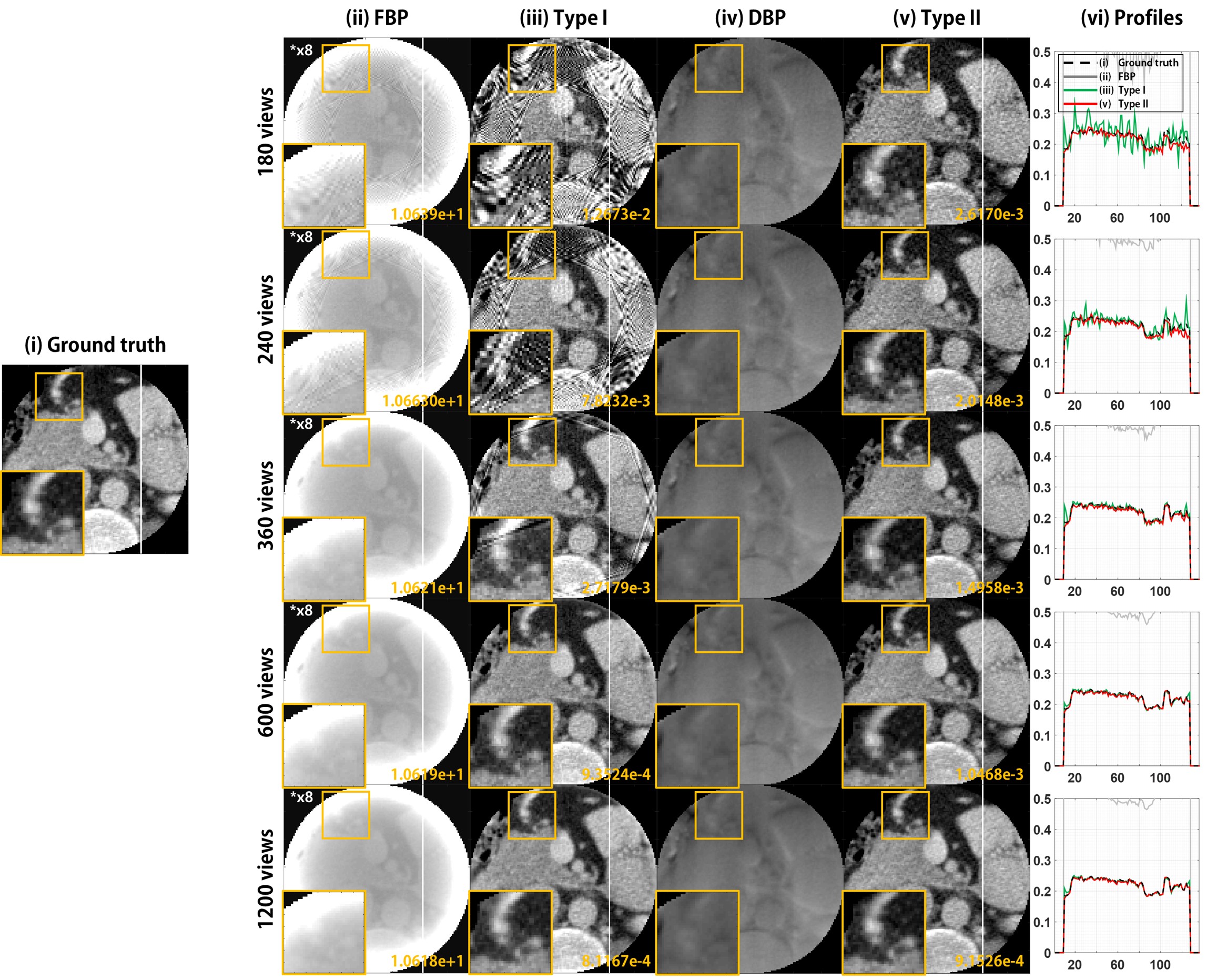} 
}
\centering
\subfigure[~\# of detectors = 240] {
	\includegraphics[width=0.3\textwidth]{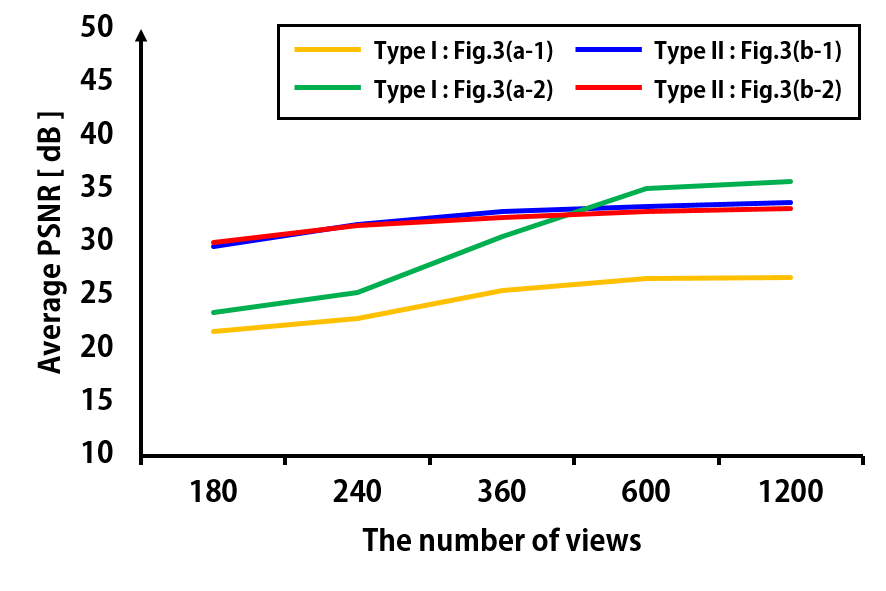} 
	\label{fig:grp_view_dct240}
}
\centering
\subfigure[~\# of detectors = 380] {
	\includegraphics[width=0.3\textwidth]{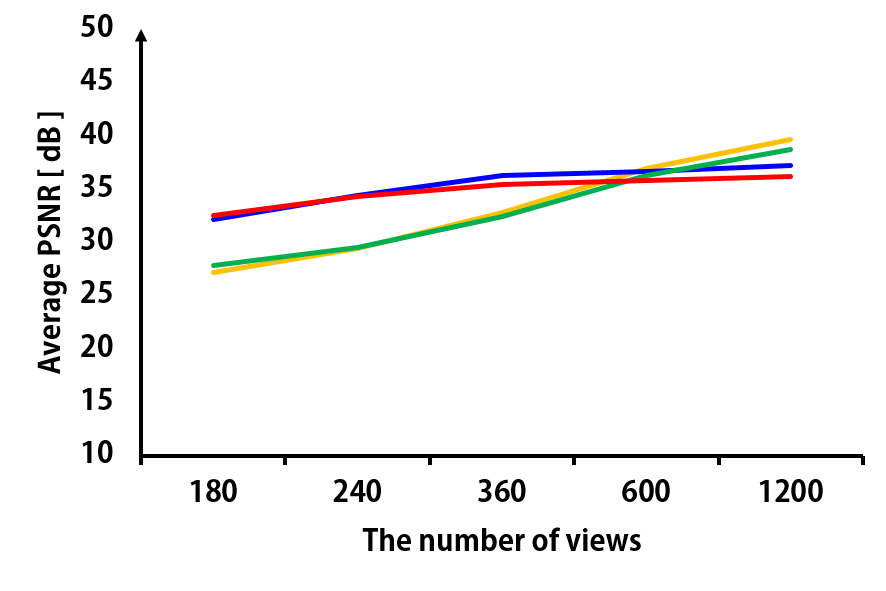} 
	\label{fig:grp_view_dct380}
}
\centering
\subfigure[~\# of detectors = 600] {
	\includegraphics[width=0.3\textwidth]{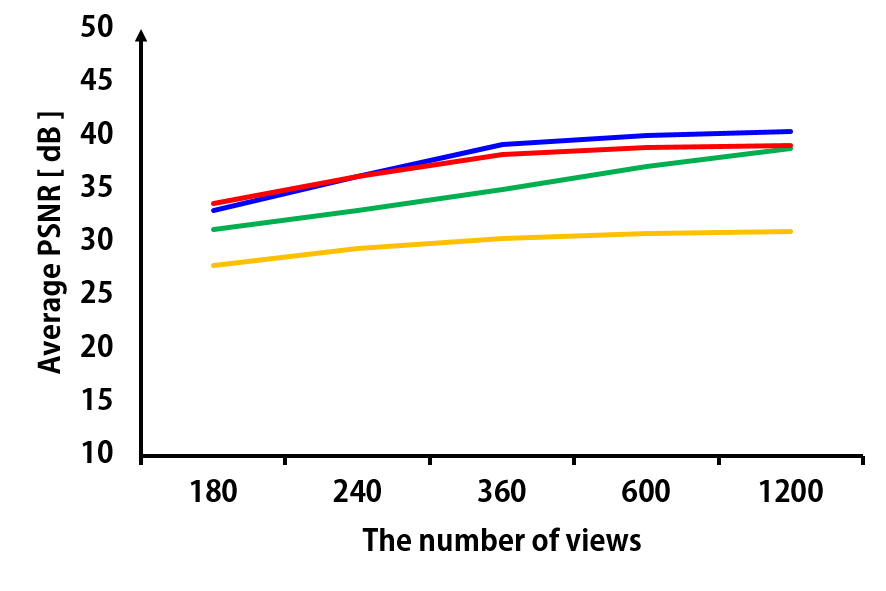} 
	\label{fig:grp_view_dct600}
}
\caption{\bf\footnotesize 
(a) Reconstruction images for various  view number for the ROI corresponding to 240 detectors.  (b-d) Average PSNR values with respect to various  projection views.
The number of detectors is (a) 240, (b) 380, and (c) 600, respectively. The NMSE values are written at the corner. A window range is (-150, 300)[HU]. The image marked $^*$x8 indicates that the window level is magnified eight times.}
\label{fig:result_axial_view_dct}
\end{figure}

If the number of views is  reduced down to 180 views, 
the FBP images as shown in Fig. \ref{fig:result_axial_view_dct}(ii) are corrupted with both cupping artifact due to truncated detector and the streaking artifact due to sparse view. 
Accordingly, the reconstructed image using Type I network  (see Fig. \ref{fig:result_axial_view_dct}(iii)) was still corrupted with the severe streaking artifacts. 
However, the reconstruction results by Type II network were less affected by the streaking artifacts, because the streaking artifacts do not appear in the DBP images as shown in Fig. \ref{fig:result_axial_view_dct}(iv).

\begin{figure}[!h]
\centering
\subfigure[ ] {
\includegraphics[width=0.7\textwidth]{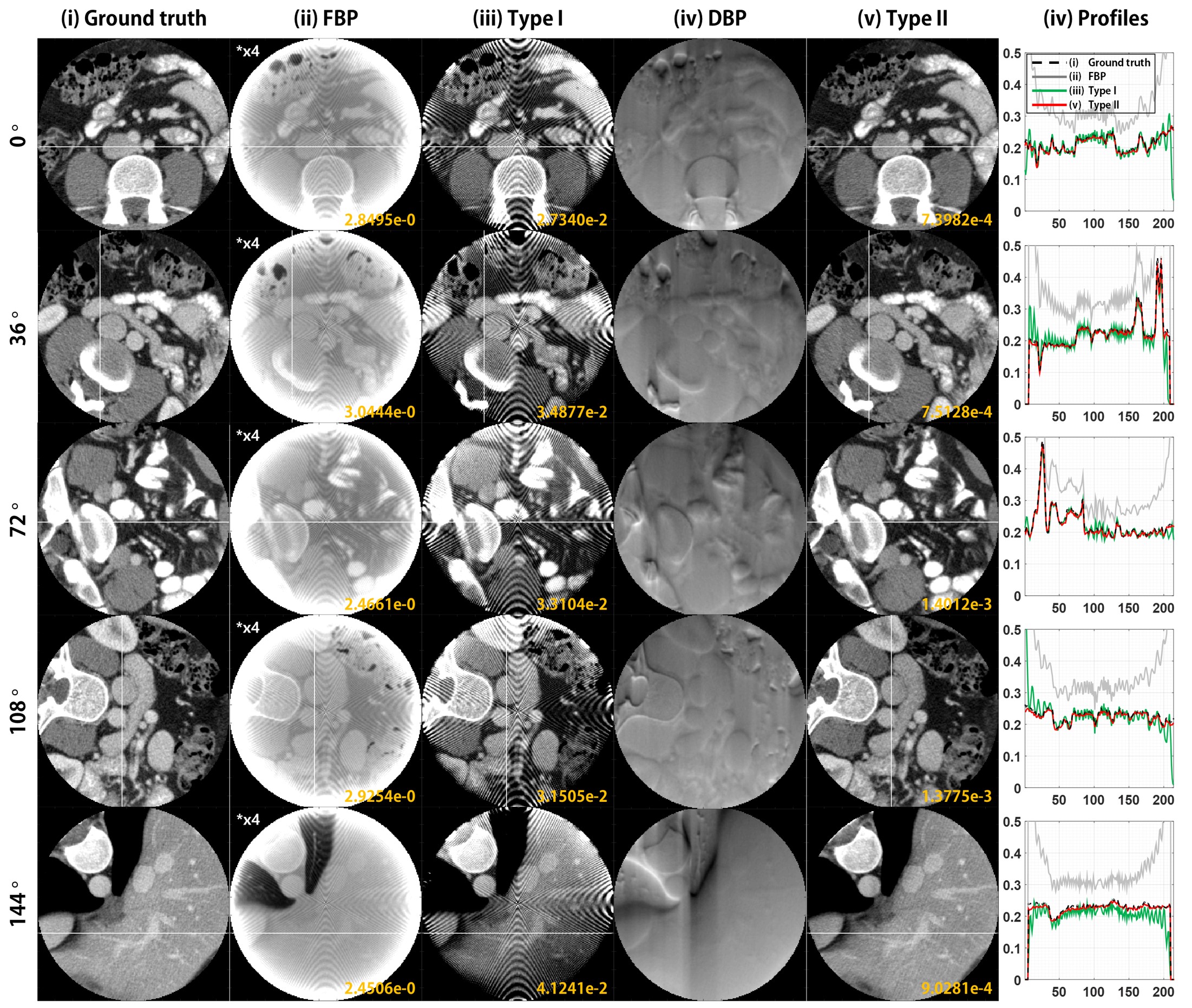} 
}
\centering
\subfigure[ ] {
\includegraphics[width=0.5\textwidth]{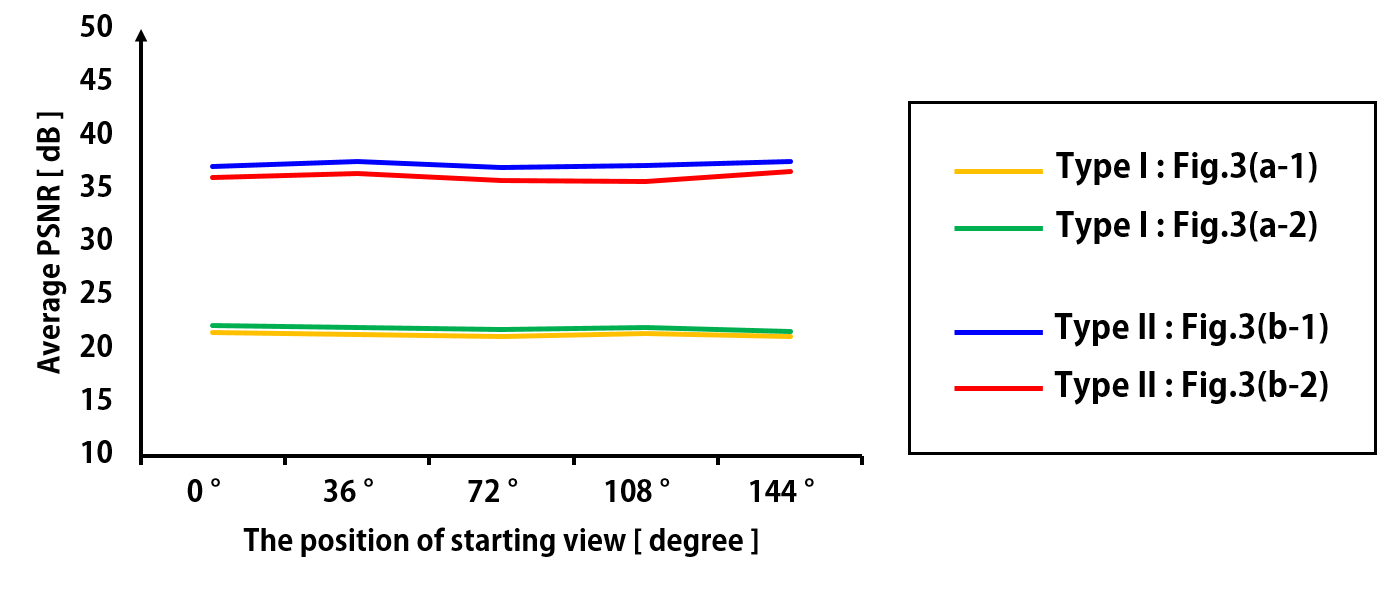} 
}
\caption{\bf\footnotesize 
(a) Reconstruction images for various starting angle for short scan imaging for the ROI corresponding to 380 detectors.  (b) Average PSNR values with respect to various starting angle.
The NMSE values are written at the corner. A window range is (-150, 300)[HU]. The image marked $^*$x4 indicates that the window level is magnified four times.}
\label{fig:result_axial_start_angle}
\end{figure}

Fig.~\ref{fig:result_axial_start_angle} show the short scan reconstruction results by Type I and Type II network.
%
Similar to the sparse view, there appeared new ripple-like patterns in the FBP images (see Fig. \ref{fig:result_axial_start_angle}(ii)). Accordingly,
 the reconstructed image using Type I network suffer from
new type of artifact patterns, whereas reconstruction results by Type II network shows the high quality reconstructed results. 
Moreover, quantitative results in Fig.~\ref{fig:result_axial_start_angle}(b) also confirm the superiority of Type II network.

\section{Discussion}

Accurate interior reconstruction  from truncated DBP data has been extensively studied
by many researchers \cite{courdurier2008solving,katsevich2012stability,jin2012interior,katsevich2012finite,yu2009compressed,ward2015interior,lee2015interior}.
Here, some kind of a priori knowledge, such as a known region, smoothness, etc., needs to be incorporated to solve the interior problem stably.
Given this, 
one may wonder  whether  the deep-learning based
methods incorporate some a priori information to make it work.
We conjecture that  the global
consistency of the reconstructed images is one of the important implicit prior information used by the neural networks.
For example, as seen from Fig.~\ref{fig:null}(a), we could generate infinite number of 1-D null space images 
by changing  $\psi(u)$. However, in order to maintain the consistency of the reconstruction across the chord lines,
the allowed choice of $\psi(u)$ may be only few, which may lead the neural network to reconstruct image uniquely.
Similarly, in \eqref{eq:formula}, to maintain the consistency across the chord lines,
the choice of the offset  $\epsilon(u)$ may be unique.
This type of consistency across chord line is more complicated than the smoothness or known interval within
each chord line, so such approaches have never been investigated to our best knowledge. Thanks to the neural networks  that can learn such complicated nonlinear mappings,  we are now able to explore
such implicit constraint. This is also the reason to explain our observation that 2-D neural
network works better than 1-D approaches.
The rigorous verification of this claim is, however, currently lacking, which may need further investigation in future research.

The Hilbert kernel has a longer decay compared to the standard ramp filter for the filtered back-projection algorithm, which should be taken consideration  in designing a neural network. 
Although the kernel size in each convolutional layer is smaller,
the total receptive field size of the kernel becomes large   after several layer convolutions and pooling layers in deep neural networks.
In fact, it was shown  in \cite{han2018framing} that the pooling layer  increases the receptive field size exponentially in contrast
to the neural network without pooling.
 In our U-Net implementation, the calculation by considering each filter kernel size and pooling layers shows that the effective
 filter size is $396 \times 396$, which is large enough to cover whole interior images; so we believe that the proposed network can effectively learn the Hilbert kernel.

Another important observation is that even for the same functional mapping, the specific weighting for each layer may not be unique, since scaled weights in one layer can be compensated by inverse scaling in the subsequent layer.
This suggests that neural network learns the dimensionless functional mapping between input and outputs, rather than specific
weighting functions with correct physical units. This may explain why the neural network can be generalized well 
for the test images with  different pixel resolution, detector pitches, etc.

Our experimental results showed that
Type I neural network gains better performance for a specific truncation severity while Type II gains better generalizability to different truncation severity. We conjecture that the specific truncation artifacts in Type I network are so distinct from the
original images that by  learning the artifact, the trained network exhibited the best performance.
However,  this type of artifacts is unique for each ROI size. As the  learning all the artifacts is difficult, this may lead to the reduced generalization capability.  Since Type II neural network learns the Hilbert transforms and the consistency across the chord lines,
such trained network may be generalized well for various acquisition parameters at the cost of reduced accuracy compared to Type I for specific truncation severity.  However,  in practical environment, the training and test environments are usually different, so we believe that the generalizability is more important.

\begin{figure}[!t]
\centering
\includegraphics[width=0.7\textwidth]{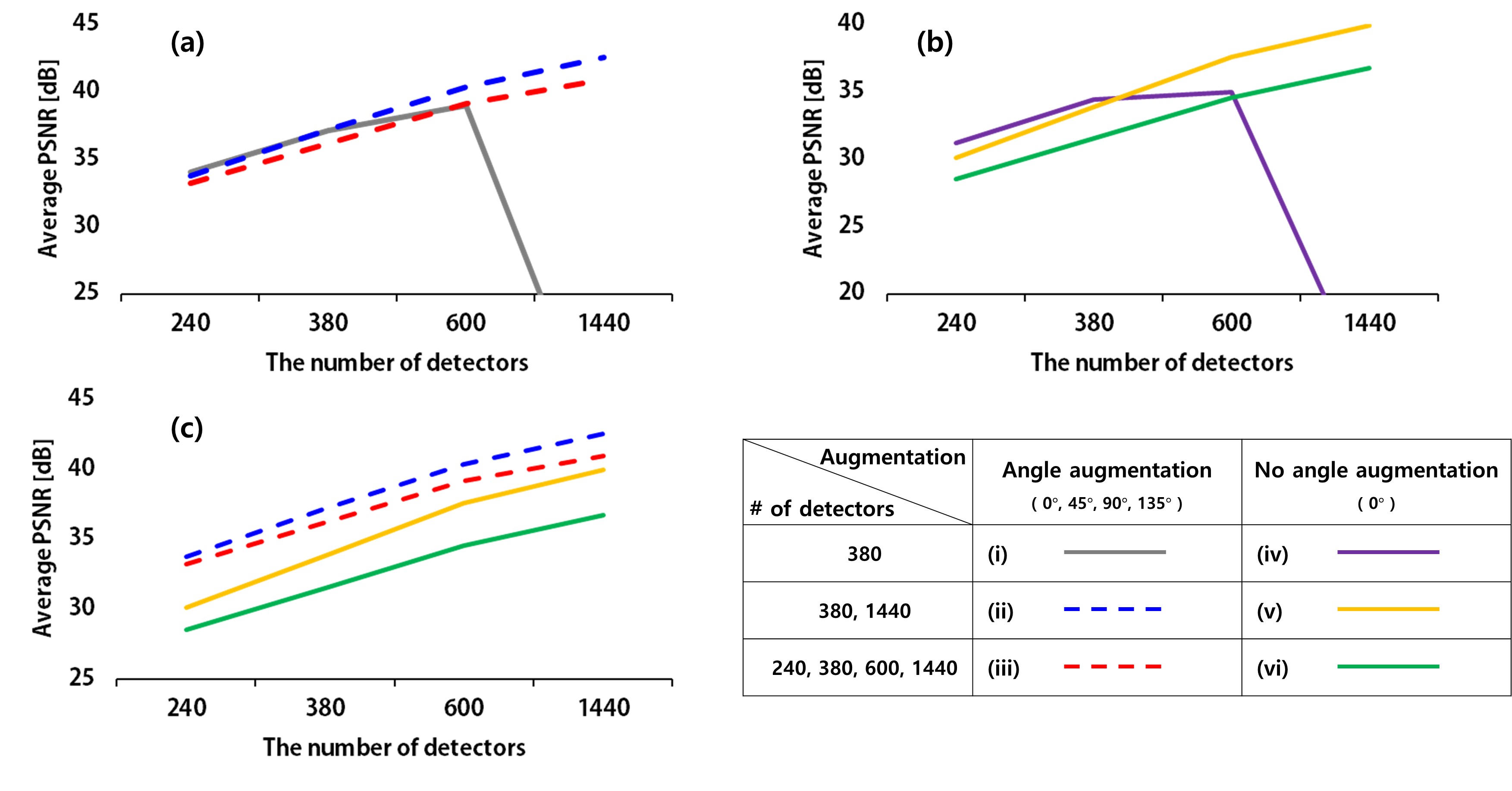} 
\caption{\bf\footnotesize Effect of detector augmentation (a) with angle augmentation, and (b) without angle augmentation.  (c) Effect of angle augmentation. }
\label{fig:augmentation}
\end{figure}

Except for the real data case, the performance of
Type II network with 240, 380, 600 and 1440 detectors were slightly worse than Type II networks with 380 and 1440 detector augmentation.
To  investigate the origin of such behavior, we have performed the additional experiment. In particular, our new experiments are designed to decouple the effects of angle and detector augmentations. First, we investigate the effect of detector augmentation. In Figure~\ref{fig:augmentation}(a), we trained three neural networks with different detector augmentation.   All the networks were angle augmented.   As shown in Figure~\ref{fig:augmentation}(a), the detector augmentation did not improve the performance of the network, and the network trained with 380 and 1440 detectors consistently outperformed the network trained with 240, 380, 600 and 1440 detectors.  Moreover, the network trained with only 380 shows best performance around 380 detector size, although the performance quickly degraded as the detector size become bigger.  
In order to investigate the potential limitation in term of network capacity,
 Figure~\ref{fig:augmentation}(b) examined  the effect of detector augmentation alone without any angle augmentation. We have observed the same phenomenon as in Figure~\ref{fig:augmentation}(a).  This strongly suggests that the detector augmentation for the Type II network reduces the specificity at the specific detector size while improving  the generalization performance. Now, we investigate the effect of the angle augmentation. As shown in Figure~\ref{fig:augmentation}(c), the angularly augmented network consistently outperforms the network without angle augmentation.
 
 We conjecture that this different behavior of data augmentation is due to the different nature of the learning in Type II network compared to Type I network. In a Type I network, the goal of the training is to learn the artifact patterns, so that the network performance is improved with training data set with more diverse detectors and angle augmentations. On the other hand, Type II network learns the truncated Hilbert transform kernels. Since the truncated Hilbert transform kernels depend on the specific truncation ratio, the effect of the augmenting data with various detector sizes may be to learn the truncated Hilbert transform kernels that works well on the  “average”; thus, with more detector augmentation, the performance degradation of the “average” kernel may be unavoidable.  In the future, however, more thorough theoretical analysis should be performed to understand this behavior.
 

In our experiment in  Fig. \ref{fig:result_axial_view_dct} and  Fig.~\ref{fig:result_axial_start_angle}, we observed that
Type I network suffers from sparse view and short scan artifacts, whereas Type II network is less prone to such artifact.
Since both networks uses the identical U-Net architecture except the input data, this appears somewhat mysterious.
In fact, the main reason for the improvement of Type II network is not from different network structure, but from better input data.
For example, 180 views in  Fig. \ref{fig:result_axial_view_dct}(iv) is in fact not too sparse, so the streaking artifacts are not very noticeable in the DBP images.  
On the other hand, due to the ramp filtering for the truncated detector measurement, the FBP image suffers from severe view aliasing artifacts  in addition to the cupping artifacts (see Fig.~\ref{fig:result_axial_view_dct}(ii)). 
Similarly, the ripple-like artifacts are observed at the FBP images (see Fig.~\ref{fig:result_axial_start_angle}(ii)) from short scan and reduced ROI due to the ramp filtering at the truncated detector boundary, whereas
such artifacts were not observed  in DBP images (see Fig.~\ref{fig:result_axial_start_angle}(iv)) due to the lack of ramp filtering.  
Recall that both Type I and Type II network were trained with full views and truncated detector measurements.  Under this condition, both networks can be trained to remove truncation artifacts. Therefore, when the FBP image like Fig. \ref{fig:result_axial_view_dct}(ii) is used as an input, Type I network can remove the truncation artifacts, but fails to remove the sparse view artifact. 
On the other hand, the DBP input image in Fig. \ref{fig:result_axial_view_dct}(iv) does not have severe streaking artifact so that the final reconstruction result by Type II network can remove the truncation artifacts without suffering from remaining sparse view artifacts. 

Given that  our Type II neural network can  freely process image with  arbitrary pixel sizes, detector size, and starting
angle for short-scan,  no matter how much
the sinogram get transversely truncated,  we believe that our network can be used also
for standard CT reconstruction.

\section{Conclusion}
\label{sec:conclusion}

In this paper, we proposed and compared two types of deep learning network for interior tomography problem.
In particular, Type I network architecture was designed to learn the artifact-free image from the analytic reconstruction, whereas
Type II network architecture was trained to learn the inverse of the finite Hilbert transform.
Due to the singularity in the artifact-corrupted images, Type I network was not well generalizable, although its performance was best at the specific ROI size used for training data.
On the other hand, the input images for  Type II network are truncated DBP data  free of singularities, making
the network generalize well under various acquisition parameters.
Numerical results showed that the proposed method significantly outperforms existing iterative methods in terms of quantitative and qualitative image quality as well as computation time.


\section*{Acknowledgement}
The authors would like to thanks Dr. Cynthia McCollough,  the Mayo Clinic, the American Association of Physicists in Medicine (AAPM), and grant EB01705 and EB01785 from the National Institute of Biomedical Imaging and Bioengineering for providing the Low-Dose CT Grand Challenge data set.
This work is supported by National Research Foundation of Korea, Grant number
NRF-2016R1A2B3008104. This work is also supported by the
R\&D Convergence Program of NST (National Research Council of Science
\& Technology) of Republic of Korea (Grant CAP-13-3-KERI).


%

\end{document}